\documentclass[10pt,twocolumn,letterpaper]{article}

\usepackage{cvpr}
\usepackage{times}
\usepackage{epsfig}
\usepackage{graphicx}
\usepackage{amsmath}
\usepackage{amssymb}

\usepackage{multirow}
\usepackage{pbox}
\usepackage{float}
\usepackage{tikz}
\usepackage{color}
\usepackage{dblfloatfix}
\usepackage{bigstrut}
\usepackage[caption=false]{subfig}

\usetikzlibrary{decorations.pathreplacing}
\usetikzlibrary{shapes.geometric}
\usetikzlibrary{matrix,positioning,arrows.meta,calc,fadings}
\usetikzlibrary{arrows}
\usetikzlibrary{fit,positioning,arrows.meta}

\newcommand{\Xt}{\tilde{X}}
\newcommand{\Yt}{\tilde{Y}}
\newcommand{\enc}{\text{Enc}}
\newcommand{\dec}{\text{Dec}}

\DeclareMathOperator{\E}{\mathbb{E}}

\usepackage[breaklinks=true,bookmarks=false]{hyperref}

\cvprfinalcopy % *** Uncomment this line for the final submission

\def\cvprPaperID{2292} % *** Enter the CVPR Paper ID here

\ifcvprfinal\fi
\begin{document}

% \title{RevGAN: A Reversible approach to unpaired image-to-image translation}
% \title{Reversibility is all you need}
% \title{Unpaired image-to-image translation with Reversible Generative Adversarial Networks}
% \title{Reversible Generative Models for Memory-efficient Image-to-image translation}
\title{Reversible GANs for Memory-efficient Image-to-Image Translation}

\author{Tycho F.A. van der Ouderaa \\
University of Amsterdam\\
%{\tt\small tycho.vanderouderaa@student.uva.nl} \\
{\tt\small tychovdo@gmail.com} \\
\and
Daniel E. Worrall \\
University of Amsterdam \\
{\tt\small d.e.worrall@uva.nl} \\
}

\maketitle
%\thispagestyle{empty}

%%%%%%%%% ABSTRACT
\begin{abstract}
The Pix2pix \cite{isola2017image} and CycleGAN \cite{zhu2017unpaired} losses have vastly improved the qualitative and quantitative visual quality of results in image-to-image translation tasks. We extend this framework by exploring approximately invertible architectures which are well suited to these losses. These architectures are approximately invertible by design and thus partially satisfy cycle-consistency before training even begins. Furthermore, since invertible architectures have constant memory complexity in depth, these models can be built arbitrarily deep. We are able to demonstrate superior quantitative output on the Cityscapes and Maps datasets at near constant memory budget.
\end{abstract}

\section{Introduction}

Computer vision was once considered to span a great many disparate problems, such as superresolution \cite{dong2014learning}, colorization \cite{Cheng15color}, denoising and inpainting \cite{Xie12denoise}, or style transfer \cite{Gatys16style}. Some of these challenges border on computer graphics (\eg style transfer), while others are more closely related to numerical problems in the sciences (\eg superresolution of medical images  \cite{tanno2017bayesian}). With the new advances of modern machine learning, many of these tasks have been unified under the term of \emph{image-to-image translation} \cite{isola2017image}.

Mathematically, given two image domains $X$ and $Y$, the task is to find or learn a mapping $F: X \to Y$, based on either paired examples $\{(x_i, y_i)\}$ or unpaired examples $\{x_i\} \cup \{y_j\}$. Let's take the example of image superresolution. Here $X$ may represent the space of low-resolution images, and $Y$ would represent the corresponding space of high-resolution images. We might equivalently seek to learn a mapping $G: Y \to X$. To learn both $F$ and $G$ it would seem sufficient to use the standard supervised learning techniques on offer, using convolutional neural networks (CNNs) for $F$ and $G$. For this, we require paired training data and a loss function $\ell$ to measure performance. In the absence of paired training data, we can instead exploit the reciprocal relationship between $F$ and $G$. Note how we expect the compositions $G \circ F \simeq \text{Id}$ and $F \circ G \simeq \text{Id}$, where $\text{Id}$ is the identity. This property is known as \emph{cycle-consistency} \cite{zhu2017unpaired}. The unpaired training objective is then to minimize $\ell(G\circ F (x), x)$ or $\ell(F\circ G (y), y)$ with respect to $F$ and $G$, across the whole training set. Notice how in both of these expressions, we never require explicit pairs $(x_i,y_i)$. Naturally, in superresolution exact equality to the identity is impossible, because the upsampling task $F$ is one-to-many, and the downsampling task $G$ is many-to-one. 

The problem with the cycle-consistency technique is that while we can insert whatever $F$ and whatever $G$ we deem appropriate into the model, we avoid making use of the fact that $F$ and $G$ are approximate inverses of one another. In this paper, we consider constructing $F$ and $G$ as approximate inverses, \emph{by design}. This is not a replacement to cycle-consistency, but an adjunct to it. A key benefit of this is that we need not have a separate $X\to Y$ and $Y\to X$ mapping, but just a single $X\to Y$ model, which we can run in reverse to approximate $Y \to X$. Furthermore, note by explicitly weight-tying the $X\to Y$ and $Y\to X$ models, we can see that training in the $X\to Y$ direction will also train the reverse $Y\to X$ direction, which does not necessarily occur with separate models. Lastly, there is also a computational benefit that invertible networks are very memory-efficient \cite{gomez2017reversible}; intermediate activations do not need to be stored to perform backpropagation. As a result, invertible networks can be built arbitrarily deep, while using a fixed memory-budget---this is relevant because recent work has suggested a trend of wider and deeper networks performing better in image generation tasks \cite{brock2018large}. Furthermore, this enables dense pixel-wise translation models to be shifted to memory-intensive arenas, such as 3D (see Section \ref{sec:brains} for our experiements on dense MRI superresolution).

Our results indicate that by using invertible networks as the central workhorse in a paired or unpaired image-to-image translation model such as Pix2pix \cite{isola2017image} or CycleGAN \cite{zhu2017unpaired}, we can not only reduce memory overhead, but also increase fidelity of the output. We demonstrate this on the Cityscapes and Maps datasets in 2D and on a diffusion tensor image MRI dataset for the 3D scenario (see Section \ref{sec:results}).

\section{Background and Related Work}
In this section, we recap the basics behind Generative Adversarial Networks (GANs), cycle-consistency, and reversible/invertible networks.

%\subsection{GANs}
\subsection{Generative Adversarial Networks (GANs)}

\emph{Generative adversarial networks} (GANs) \cite{goodfellow2014generative} enjoy huge success in tasks such as image generation \cite{brock2018large}, image interpolation \cite{karras17progressive}, and image re-editing \cite{Perarnau16editing}. They consist of two components, a generator $F: Z \to Y$ mapping random noise $z\in Z$ to images $y\in Y$ and a discriminator $D: Y \to [0,1]$ mapping images $y\in Y$ to probabilities. Given a set of training images $\{y_1, y_2, ...\}$, the generator produces `fake' images $y_* = F(z), z\sim p(z)$, where $p(z)$ is a simple distribution such as a standard Gaussian, and the discriminator tries to predict the probability that the image was from the true image distribution. For training, an \emph{adversarial loss} $L_{\text{GAN}}$ is defined:
\begin{align}
L_{\text{GAN}}(F, D) =\E_{y} \log D(y) +\E_{z} \log (1 - D(F(z)))
\end{align}

This loss is trained using a minimax regime where intuitively we encourage the generator to fool the discriminator, while also training the discriminator to guess whether the generator created an image or not. Mathematically this game \cite{goodfellow2014generative} is
\begin{align}
    F^* = \arg \min_F \max_{D} L_{\text{GAN}}(F, D).
\end{align}
At test time, the discriminator is discarded and the trained generator is used to hallucinate fake images from the same distribution \cite{Arora17distribution} as the training set. The generator can be conditioned on an input image as well. This setup is called a \emph{conditional GAN} \cite{Mirza14cgan}.

\subsection{Image-to-Image Translation}
In a standard (paired) image-to-image translation problem \cite{isola2017image}, we seek to learn the mapping $F: X \to Y$, where $X$ and $Y$ are corresponding spaces of images. It is natural to model $F$ with a convolutional neural network (CNN). To train this CNN we minimize a loss function 
\begin{align}
    L(F) = \frac{1}{n}\sum_{i=1}^n \ell(F(x_i),y_i)
\end{align} 
where $\ell$ is a loss function defined in the pixel-space between the prediction $F(x_i)$ and the target $y_i$. Traditional image-to-image translation tasks relying on pixel-level loss functions are hampered by the fact that these losses do not typically account for inter-pixel correlations \cite{Zhao17losses}, for instance, $L_1$-losses treat each pixel as independent. Instead, since GANs do not apply the loss per-pixel, they can account for these inter-pixel correlational structures. GANs can be co-opted for image-to-image translation by adding the adversarial loss on top of a standard pixel-level $L_1$ loss function. This was first performed in the Pix2pix model \cite{isola2017image}, which is for paired image-to-image translation problems. Pix2pix replaces $F$ with a conditional generator $F: X \times Z \rightarrow Y$, where $Z$ is the domain of the random noise; although, in practice, we usually ignore the additional noise input \cite{zhu2017unpaired}. The model combines a $L_1$-loss that enforces the model to map images to the paired translations in a supervised manner with an adversarial loss that enforces the model to adopt the style of the target domain. The loss is
\begin{align}
    F^* = \arg \min_F \max_D L_{\text{cGAN}}(F,D) + \lambda L_{L1}(F) \label{eq:pix2pix}
\end{align}
where 
\begin{align}
	L_{L1}(F) &= \mathbb{E}_{x,y} \|y - F(x)\|_1 \\
	L_{\text{cGAN}}(F,D) &= \E_{x} \left [ \log D(x) + \log (1 - D(F(x)) \right ].
\end{align}
$\lambda$ is a tuneable hyperparameter typically set in the range $10-100$ \cite{isola2017image}.

\subsection{Cycle-consistency}
The CycleGAN model was proposed as an alternative to Pix2pix for unpaired domains \cite{zhu2017unpaired}. The model uses two generators $F$ and $G$ for the respective mappings between the two domains $X$ and $Y$ (so, $F: X \rightarrow Y$ and $G: Y \rightarrow X$), and two discriminators $D_X: X \to [0,1] $ and $D_Y : Y \to [0, 1]$ trained to distinguish real and generated images in both domains. Since there are no image pairings between domains, we cannot invoke the Pix2pix loss and instead CycleGAN uses a separate \textit{cycle-consistency loss} that penalizes the distances $L _{\text{cycle}}(G,F,x) = \|G \circ F (x) - x\|_1$ and $L_{\text{cycle}}(F,G,y) =  \|F \circ G (y) - y\|_1$ across the training set. This encourages that the mappings $F$ and $G$ are loose inverses of one another. This allows the model to train on unpaired data. The total loss is
\begin{align}
	L_{\text{cycleGAN}} &= L_{\text{cGAN}}(F,D_Y) + L_{\text{cGAN}}(G,D_X) \notag \\
	&\quad+ \mathbb{E}_x L_{\text{cycle}}(G,F,x) + \mathbb{E}_y L_{\text{cycle}}(F,G,y). \label{eq:cycleGAN}
\end{align}

Given that $F$ and $G$ are loose inverses of one another, it seems wasteful to use separate models to model each. In this paper, we model $F$ and $G$ as approximate inverses of one another. For this, we make use of the new area of invertible neural networks.

\subsection{Invertible Neural Networks (INNs)}
In recent years, several studies have proposed invertible neural networks (INNs) in the context of normalizing flow-based methods \cite{rezende2015variational} \cite{kingma2016improved}. It has been shown that INNs are capable of generating high quality images \cite{kingma2018glow}, perform image classification without information loss in the hidden layers \cite{jacobsen2018revnet} and analyzing inverse problems \cite{ardizzone2018analyzing}. Most of the work on INNs, including this study, heavily relies upon the transformations introduced in NICE \cite{dinh2014nice} later extended in RealNVP \cite{dinh2016density}. Although INNs share interesting properties they remain relatively unexplored.

\tikzset{fontscale/.style = {font=\relsize{#1}}}
\tikzstyle{stateTransition}=[-stealth, thick]

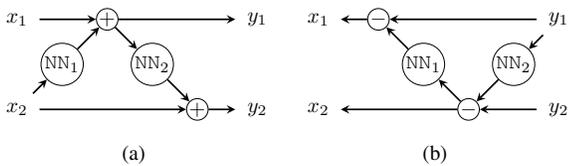
\begin{figure}[t]
\centering
\subfloat[]{
\resizebox{0.47\linewidth}{!}{
\begin{tikzpicture}
 \node[circle] (x1) at (0,0)  {$x_1$};
 \node[circle] (x2) at (0,-1.5) {$x_2$};
 \node[circle] (y1) at (4,0)  {$y_1$};
 \node[circle] (y2) at (4,-1.5) {$y_2$};
 \node[circle,draw=black, inner sep=0pt,minimum size=20pt] (F) at (0.75,-0.75)  {$\texttt{NN}_1$};
 \node[circle,draw=black, inner sep=0pt,minimum size=20pt] (G) at (2.25,-0.75)  {$\texttt{NN}_2$};
 \node[circle,draw=black, inner sep=0pt,minimum size=10pt] (p1) at (1.5, 0)  {$+$};
 \node[circle,draw=black, inner sep=0pt,minimum size=10pt] (p2) at (3, -1.5)  {$+$};
 \draw[stateTransition] (x1) edge (p1);
 \draw[stateTransition] (p1) edge (y1);
 \draw[stateTransition] (x2) edge (p2);
 \draw[stateTransition] (p2) edge (y2);
 \draw[stateTransition] (x2) edge (F);
 \draw[stateTransition] (F) edge (p1);
 \draw[stateTransition] (p1) edge (G);
 \draw[stateTransition] (G) edge (p2);
\end{tikzpicture}
}}
\subfloat[]{
\resizebox{0.47\linewidth}{!}{
\begin{tikzpicture}
  \node[circle] (x1) at (0,0)  {$x_1$};
  \node[circle] (x2) at (0,-1.5) {$x_2$};
  \node[circle] (y1) at (4,0)  {$y_1$};
  \node[circle] (y2) at (4,-1.5) {$y_2$};
  \node[circle,draw=black, inner sep=0pt,minimum size=20pt] (F) at (1.75,-0.75)  {$\texttt{NN}_1$};
  \node[circle,draw=black, inner sep=0pt,minimum size=20pt] (G) at (3.25,-0.75)  {$\texttt{NN}_2$};
  \node[circle,draw=black, inner sep=0pt,minimum size=10pt] (m1) at (1, 0)  {$-$};
  \node[circle,draw=black, inner sep=0pt,minimum size=10pt] (m2) at (2.5, -1.5)  {$-$};
  \draw[stateTransition] (y1) edge (m1);
  \draw[stateTransition] (m1) edge (x1);
  \draw[stateTransition] (y2) edge (m2);
  \draw[stateTransition] (m2) edge (x2);
  \draw[stateTransition] (y1) edge (G);
  \draw[stateTransition] (G) edge (m2);
  \draw[stateTransition] (m2) edge (F);
  \draw[stateTransition] (F) edge (m1);
\end{tikzpicture}
}}
\medskip
\caption{Illustration of forward (a) and backward (b) pass of a residual block in a reversible residual layer. Note, functions $\texttt{NN}_1$ and $\texttt{NN}_2$ need not be invertible, hence the layer is very flexible and in practical terms very easy to implement.}
\label{fig:additive_coupling}
\end{figure}

\paragraph{Additive Coupling}
In our model, we obtain an invertible residual layer, as used in \cite{gomez2017reversible}, using a technique called \textit{additive coupling} \cite{dinh2014nice}: first we split an input $x$ (typically over the channel dimension) into $(x_1, x_2)$ and then transform them using arbitrary complex functions $\texttt{NN}_1$ and $\texttt{NN}_2$ (such as a ReLU-MLPs) in the form (left):

\begin{align}
y_1 &= x_1 + \texttt{NN}_1(x_2) \qquad && x_1 = y_1 - \texttt{NN}_1(x_2)\\
y_2 &= x_2 + \texttt{NN}_2(y_1) \qquad && x_2 = y_2 - \texttt{NN}_2(y_1).
\end{align}

The inverse mappings can be seen on the right. Figure \ref{fig:additive_coupling} shows a schematic of these equations.

\paragraph{Memory efficiency}
Interestingly, invertible residual layers are very memory-efficient because intermediate activations do not have to be stored to perform backpropagation \cite{gomez2017reversible}. During the backward pass, input activations that are required for gradient calculations can be (re-)computed from the output activations because the inverse function is accessible. This results in a constant spatial complexity ($\mathcal{O}(1)$) in terms of layer depth (see Table \ref{tab:mem_complexity}).

\begin{table}[b]
\centering
\resizebox{\linewidth}{!}{
\begin{tabular}{l l l}
\hline
Technique & \pbox{2.7cm}{Spatial Complexity\\(Activations)} & \pbox{2cm}{Computational\\Complexity} \\
\hline
Naive & $\mathcal{O}(L)$ & $\mathcal{O}(L)$ \\
Checkpointing \cite{martens2012training} & $\mathcal{O}(\sqrt{L})$ & $\mathcal{O}(L)$\\
\pbox{4cm}{Recursive} \cite{chen2016training} & $\mathcal{O}(\log L)$ & $\mathcal{O}(L \log L)$ \\
Additive Coupling \cite{gomez2017reversible} & $\mathcal{O}(1)$ & $\mathcal{O}(L)$ \\ \hline
\end{tabular}
}
\medskip
\caption{Comparison of Spatial and Computational Complexity copied from \cite{gomez2017reversible}. $L$ denotes number of residual layers. Notice how the spatial complexity of additive coupling is $\mathcal{O}(1)$ versus $\mathcal{O}(L)$ for a naive implementation.}
\label{tab:mem_complexity}
\end{table}

\tikzstyle{h_enc} = [
    draw,
    trapezium,
    trapezium stretches body,
    trapezium angle=75,
    shape border rotate=280,
    minimum height=1 cm,
    minimum width=1 cm,
    node distance=1.05 cm,
    inner sep=0pt,
    ]
\tikzstyle{h_dec} = [
    draw,
    trapezium,
    trapezium stretches body,
    trapezium angle=75,
    shape border rotate=90,
    minimum height=1 cm,
    minimum width=1 cm,
    node distance=1.05 cm,
    inner sep=0pt,
    ]
\tikzstyle{h_block} = [
    draw,
    minimum width=1 cm,
    minimum height=1 cm,
    node distance=1.5 cm
    ]
\tikzstyle{h_arr} = [
           ->,
           shorten <=0.2em,
           shorten >=0.2em]

\tikzstyle{block} = [draw, fill=white, rectangle, minimum height=4em, minimum width=4em]
\tikzstyle{tinyblock} = [draw, fill=white, rectangle, minimum height=1.8em, minimum width=1.8em]
\tikzstyle{pinstyle} = [pin edge={to-,thin,black}]
\tikzstyle{arr} = [
           ->,
           shorten <=0.5em,
           shorten >=0.5em]
           
\begin{figure*}[!t]
    \centering
    {
        \resizebox{0.8\linewidth}{!}{
            \begin{tikzpicture}[->,>=stealth',auto,node distance=3cm,]
    \node [block, pin={[pinstyle]above:$D_X$}, node distance=3cm] (block_x) {$X$};
    \node [block, right of=block_x , node distance=3cm] (block_x2) {$\tilde{X}$};
    \node [block, right of=block_x2, node distance=3cm] (block_y2) {$\tilde{Y}$};
    \node [block, right of=block_y2, pin={[pinstyle]above:$D_Y$}, node distance=3cm] (block_y) {$Y$};
    \path (block_x.east) edge [arr, bend left=45] node {$\text{Enc}_X$} (block_x2.west);
    \path (block_x2.west) edge [arr, bend left=45] node {$\text{Dec}_X$} (block_x.east);
    \path (block_x2.east) edge [arr, bend left=45] node {$C$} (block_y2.west);
    \path (block_y2.west) edge [arr, bend left=45] node {$C^{-1}$} (block_x2.east);
    \path (block_y2.east) edge [arr, bend left=45] node {$\text{Dec}_Y$} (block_y.west);
    \path (block_y.west) edge [arr, bend left=45] node { $\text{Enc}_Y$} (block_y2.east);
            \end{tikzpicture}
        }
    }
    \medskip
    \caption{Schematic of our RevGAN model. Between the low-dimensional image spaces $X$ and $Y$ and their corresponding high-dimensional feature spaces $\Xt$ and $\Yt$ we place non-invertible encoder and decoder networks $\text{Enc}_X, \text{Dec}_X, \text{Enc}_Y$ and $\text{Dec}_Y$. The feature spaces $\Xt$ and $\Yt$ are of the same dimension, and between them we place an invertible core network $C$. We also attach to each image space, $X$ and $Y$ a domain-specific discriminator, which is used for training with the adversarial loss.}
    \label{fig:revgan-model}
\end{figure*}
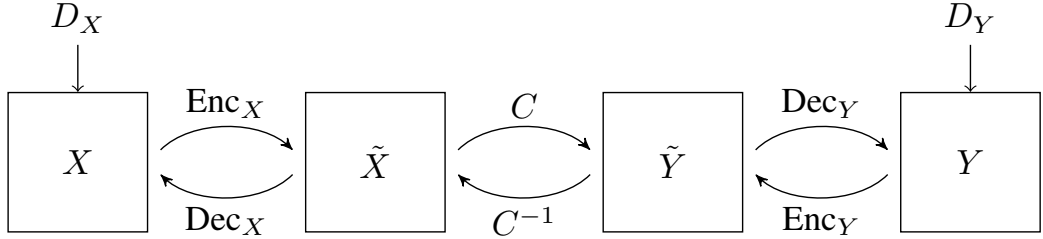

\section{Method}
Our goal is to create a memory-efficient image-to-image translation model, which is approximately invertible \emph{by design}. Below we describe the basic outline of our approach of how to create an approximately-invertible model, which can be inserted into the existing Pix2pix and CycleGAN frameworks. We call our model \emph{RevGAN}.

\paragraph{Lifting and Projection}

In general, image-to-image translation tasks are not one-to-one. As such, a fully invertible treatment is undesirable, and sometimes in the case of dimensionality mismatches, impossible. Furthermore, it appears that the high-dimensional, overcomplete representations used by most modern networks lead to faster training \cite{ochs18lifting} and better all-round performance \cite{brock2018large}. We therefore split the forward $F:X\to Y$ and backward $G: Y \to X$ mappings into three components. With each domain, $X$ and $Y$, we associate a high-dimensional feature space $\Xt$ and $\Yt$, respectively. There are individual, non-invertible mappings between each image space and its corresponding high-dimensional feature-space; for example, for image space $X$ we have $\enc_X: X \to \Xt$ and $\dec_X: \Xt \to X$. $\enc_X$ \emph{lifts} the image into a higher dimensionality space and $\dec_X$ \emph{projects} the image back down into the low-dimensional image space. We have used the terms \emph{encode} and $\emph{decode}$ in place of `lifting' and `projection' to stay in line with the deep learning literature. 

\paragraph{Invertible core}

Between the feature spaces, we then place an \emph{invertible core} $C: \Xt \to \Yt$, so the full mappings are
\begin{align}
    F &= \dec_Y \circ C \circ \enc_X \\
    G &= \dec_X \circ C^{-1} \circ \enc_Y.
\end{align}
For the invertible cores we use invertible residual networks based on additive coupling as in \cite{gomez2017reversible}. The full mappings $F$ and $G$ will only truly be inverses if $\enc_X \circ \dec_X = \text{Id}$ and $\enc_Y \circ \dec_Y = \text{Id}$, which cannot be true, since the image spaces are lower dimensional than the feature spaces. Instead, these units are trained to be approximately invertible pairs via the end-to-end cycle-consistency loss. Since the encoder and decoder are not necessarily invertible they can consist of non-invertible operations, such as pooling and strided convolutions. 

Because both the core $C$ and its inverse $C^{-1}$ are differentiable functions (with shared parameters), both functions can both occur in the forward-propagation pass and are trained simultaneously. Indeed, training $C$ will also train $C^{-1}$ and vice versa. The invertible core essentially weight-ties in the $X \to Y$ and $Y \to X$ directions.

Given that we use the cycle-consistency loss it may be asked, why do we go to the trouble of including an invertible network? The reason is two-fold: firstly, while image-to-image translation is not a bijective task, it is close to bijective. A lot of the visual information in an image $x$ should reappear in its paired image $y$, and by symmetry a lot of the visual information in the image $y$ should appear in $x$. It thus seems sensible that the networks $F$ and $G$ should be at least initialized, if not loosely coupled to be weak inverses of one another. If the constraint of bijection is too high, then the models can learn to diverge from bijection via the non-invertible encoders and decoders. Secondly, there is a potent argument for using memory efficient networks in these memory-expensive, dense, pixel-wise regression tasks. The use of two separate reversible networks is indeed a possibility for both $F$ and $G$. These would both have constant memory complexity in depth. Rather than having two networks, we can further reduce the memory budget by a rough factor of about two by exploiting the loose bijective property of the task, sharing the $X \to Y$ and $Y \to X$ models.

% \subsection{Reversible Model for Paired Data}
% \subsection{Reversible Pix2pix}
\paragraph{Paired RevGAN}
We train our paired, reversible, image-to-image translation model, using the standard Pix2pix loss functions of Equation \ref{eq:pix2pix} from \cite{isola2017image}, applied in the $X\to Y$ and $Y\to X$ directions:
\begin{align}
L_{\text{RevGANpaired}} &= \lambda(L_\text{L1}(F) + L_\text{L1}(G)) \notag \\
&\quad + L_{\text{cGAN}}(F,D_Y) + L_{\text{cGAN}}(G,D_X)
\end{align}
We also experimented with extra input noise for the conditional GAN, but found it not to help.

% Finally, a $\lambda$ parameter determines the relative importance between the L1 and the adversarial loss, resulting in the following combined loss:
% \begin{align*}
% \text{L}(G, F, D_X, D_Y)
% &= \text{L}_{\text{GAN}}(G, D_Y) \\
% &+ \text{L}_{\text{GAN}}(F, D_X) \\
% &+ \lambda \text{L}_{\text{L1}}(G, F)
% \end{align*}

% \subsection{Reversible Model for Unpaired Data}
% \subsection{Reversible CycleGAN}
\paragraph{Unpaired RevGAN} 

For unpaired RevGAN, we adapt the loss functions of the CycleGAN model \cite{zhu2017unpaired}, by replacing the $L_1$ loss with a cycle-consistency loss, so the total objective is:

\begin{align}
	&L_{\text{RevGANunpaired}} = L_{\text{cGAN}}(F,D_Y) + L_{\text{cGAN}}(G,D_X) \notag \\
	&\qquad + \mathbb{E}_x L_{\text{cycle}}(G,F,x) + \mathbb{E}_y L_{\text{cycle}}(F,G,y). \label{eq:cycleRevGAN}
\end{align}
 
% resulting in a combined loss function:
% \begin{align*}
% \text{L}(G, F, D_X, D_Y)
% &= \text{L}_{\text{GAN}}(G, D_Y) \\
% &+ \text{L}_{\text{GAN}}(F, D_X) \\
% &+ \lambda \text{L}_{\text{CYC}}(G, F)
% \end{align*}
% 
% where the relative importance between the two objectives can be controlled using the $\lambda$ parameter. Finally, the goal is to find the best mappings $G*$ and $F*$ by solving for the following mini-max game:
% \begin{align*}
% G*, F* = \arg \min_{G, F} \max_{D_X, D_Y} \text{L}(G, F, D_X, D_Y)
% \end{align*}

\section{Implementation and datasets}
The model we describe is very general and so below we explain in more detail the specifics of how to implement our paired and unpaired RevGAN models. We present 2D and 3D versions of the reversible models.

\subsection{Implementation}
\paragraph{Network Architectures}
We use two main varieties of architecture. On the 2D problems, we modify the ResNet from \cite{zhu2017unpaired}, by replacing the inner convolutions with a reversible core. The core consists of 6 or 9 reversible residual layers, dependent on the dataset---we use 6 reversible residual layers for the core on $128\times128$ (\textit{Cityscapes}) data and 9 reversible residual layers on $256 \times 256$ (\textit{Maps}) data. A more detailed description of the network architectures can be found in the supplementary material. In 3D, we use an architecture based on the SRCNN of \cite{dong2014learning} (more details in supplementary material).

\paragraph{Training details}
All model parameters were initialized from a Gaussian distribution with mean 0 and standard deviation 0.02. For training we used the Adam optimizer \cite{kingma2014adam} with a learning rate of 0.0002 (and $\beta_1=0.5, \beta_2=0.999$). We keep the learning rate fixed for the first 100 epochs and then linearly decay the learning rate to zero over the next 100 epochs, for the 2D models. The 3D models are trained with a fixed learning rate for 20 epochs. We used a $\lambda$ factor of $10$ for the unpaired models and a $\lambda$ factor of $100$ for the paired models.

\begin{figure*}[!t]
\resizebox{\linewidth}{!}{
\begin{tabular}{c | c c | c c }
Input & CycleGAN & \textbf{Unpaired RevGAN (ours)} & Pix2pix & \textbf{Paired RevGAN (ours)}  \\
\includegraphics[height=1.9cm,width=3.8cm]{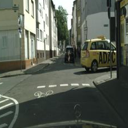}&
\includegraphics[height=1.9cm,width=3.8cm]{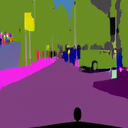} &
\includegraphics[height=1.9cm,width=3.8cm]{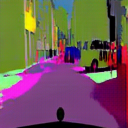} &
\includegraphics[height=1.9cm,width=3.8cm]{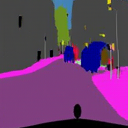} &
\includegraphics[height=1.9cm,width=3.8cm]{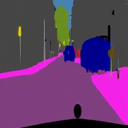} \\ 

\includegraphics[height=1.9cm,width=3.8cm]{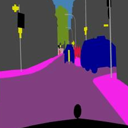} &
\includegraphics[height=1.9cm,width=3.8cm]{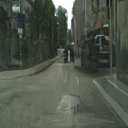} &
\includegraphics[height=1.9cm,width=3.8cm]{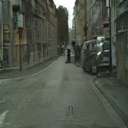} &
\includegraphics[height=1.9cm,width=3.8cm]{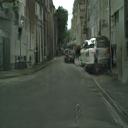} &
\includegraphics[height=1.9cm,width=3.8cm]{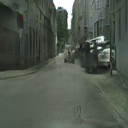}
\end{tabular}
}
\medskip
\caption{Test set image mappings on the \textit{Cityscapes} dataset for the CycleGAN and Pix2pix models, compared to our reversible variants. \textsc{Top}: The photo$\rightarrow$label mapping. \textsc{Bottom}: The label$\rightarrow$photo mapping. Notice how in the greatest improvement is between the CycleGAN and our unpaired RevGAN variant; whereas, both the Pix2pix and paired RevGAN models are of comparative visual fidelity. More results can be found in the supplementary material.}
\label{fig:cityscapes_images}
\end{figure*}

\begin{table*}[!b]
\centering
\resizebox{\linewidth}{!}{\begin{tabular}{l r r l l l c l l l}
\hline
\multicolumn{1}{c}{\multirow{2}{*}{\pbox{2cm}{Model}}} & 
\multicolumn{1}{c}{\multirow{2}{*}{\pbox{2cm}{Width}}} & 
\multicolumn{1}{c}{\multirow{2}{*}{\pbox{2cm}{Params}}}
& \multicolumn{3}{c}{photo$\rightarrow$label}
&
& \multicolumn{3}{c}{label$\rightarrow$photo}
\\
\cline{4-6} \cline{8-10}
& & & Per-pixel acc. & Per-class acc. & Class IOU & & Per-pixel acc. & Per-class acc. & Class IOU\\ \hline
CycleGAN (baseline)$^\dagger$ & 32 & 3.9 M &  0.60 & \textbf{0.27} & \textbf{0.19} && 0.42 & 0.15 & 0.10 \\
Unpaired RevGAN & 32 & 1.3 M & 0.52 & 0.21 & 0.14 && 0.36 & 0.14 & 0.09 \\
Unpaired RevGAN$^{\dagger}$  & 56  & 3.9 M & \textbf{0.66} & 0.25 & 0.18 && \textbf{0.65} & \textbf{0.24} & \textbf{0.17} \\
%Unpaired RevGAN & 64 & 5.1 M & 0.67 & 0.26 & 0.19 && 0.51 & 0.20 & 0.13 \\ \hline
\hline
Pix2pix (baseline)$^{\dagger}$ & 32 & 3.9 M & \textbf{0.82} & 0.43 & 0.32 && \textbf{0.61} & \textbf{0.22} & \textbf{0.16} \\
Paired RevGAN & 32   & 1.3 M & 0.81 & 0.41 & 0.31 && 0.57 & 0.20 & 0.15 \\
Paired RevGAN$^{\dagger}$ & 56   & 3.9 M & \textbf{0.82} & \textbf{0.44} & \textbf{0.33} && 0.60 & 0.21 & \textbf{0.16}\\
%Paired RevGAN & 64   & 5.1 M & 0.82 & 0.46 & 0.34 && 0.60 & 0.21 & 0.16 \\ \hline
\end{tabular}
}
\medskip
\caption{\textsc{Center} Classification scores on \textit{Cityscapes} photo$\rightarrow$label. \textsc{Right} FCN-scores on \textit{Cityscapes} label$\rightarrow$photo. \textsc{Top} Unpaired models. \textsc{Bottom} Paired models. Bold numbers indicate where the best model in that section. Notice that in the sections where the baseline beats our model, the differences in values are only very small. $\dagger$ Parameter matched architectures}
\label{tab:cityscapes_results}
\end{table*}

\subsection{Datasets}
We run tests on two 2D datasets and one 3D dataset. All three datasets have paired $X$ and $Y$ domain images, and we can thus extract quantitative evaluations of image fidelity.

\paragraph{Cityscapes}
The Cityscapes dataset \cite{cordts2016cityscapes} is comprised of urban street scenes with high quality pixel-level annotations. For comparison purposes, we used the same 2975 image pairs as used in \cite{zhu2017unpaired} for training and the validation set for testing. All images were downsampled to $128 \times 128$.

For evaluation, we adopt commonly used semantic segmentation metrics: \textit{per-pixel accuracy}, \textit{per-class accuracy} and \textit{class intersection-over-union}. The outputs of photo$\rightarrow$label mappings can directly be evaluated. For the reverse mapping, label$\rightarrow$photo, we use the \textit{FCN-Score} \cite{zhu2017unpaired}, by first passing our generated images through a FCN-8s semantic segmentation model \cite{long2015fully} separately trained on the same segmentation task. We then measure the quality of the obtained segmentation masks using the same classification metrics. The intuition behind this (pseudo-)metric is that the segmentation model should perform well if images generated by the image-to-image translation model are of high quality.

\begin{figure*}[!t]
\resizebox{\linewidth}{!}{
\begin{tabular}{c | c c | c c | c}
Input & CycleGAN & \textbf{Unpaired RevGAN (ours)} & Pix2pix & \textbf{Paired RevGAN (ours)} & Ground-truth \\
\includegraphics[width=0.17\linewidth]{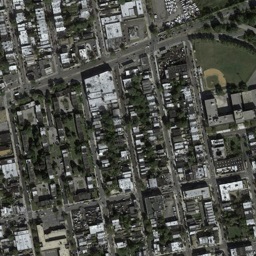} &
\includegraphics[width=0.17\linewidth]{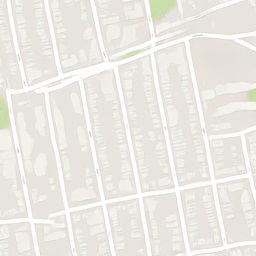} &
\includegraphics[width=0.17\linewidth]{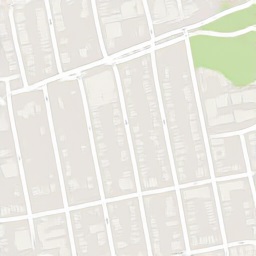} &
\includegraphics[width=0.17\linewidth]{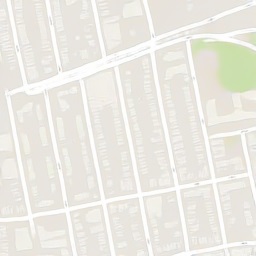} &
\includegraphics[width=0.17\linewidth]{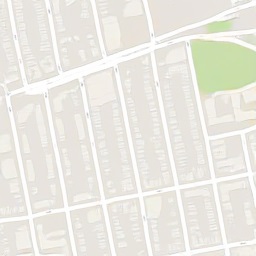} &
\includegraphics[width=0.17\linewidth]{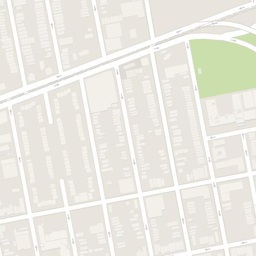} \\
\includegraphics[width=0.17\linewidth]{images/maps_images_jpg/1_maps_cycle_real_B.jpg} &
\includegraphics[width=0.17\linewidth]{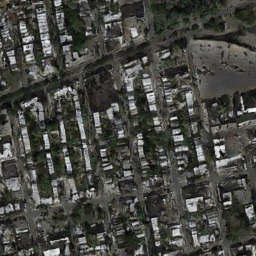} &
\includegraphics[width=0.17\linewidth]{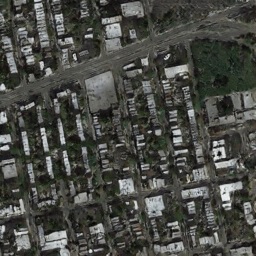} &
\includegraphics[width=0.17\linewidth]{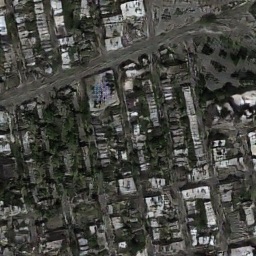} &
\includegraphics[width=0.17\linewidth]{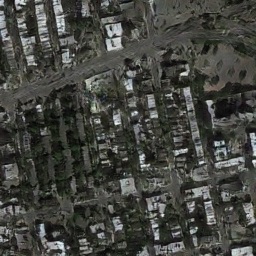} &
\includegraphics[width=0.17\linewidth]{images/maps_images_jpg/1_maps_cycle_real_A.jpg}\\
\end{tabular}
}
\medskip
\caption{Test set image mappings on the \textit{Maps} dataset. We see from this panel of images that there is no obvious degradation in the quality of the translated images between the baselines (Pix2pix and CycleGAN) and the reversible variants.}
\label{fig:maps_images}
\end{figure*}

\paragraph{Maps} The Maps dataset contains 1096 training images and an equally sized test set carefully scraped from Google Maps in and around New York City by \cite{isola2017image}. The images in this dataset are downsampled to $256 
\times 256$.

We evaluate the outputs with several commonly used metrics for image-quality: \textit{mean absolute error} (MAE), \textit{peak signal-to-noise ratio} (PSNR) and the \textit{structural similarity index} (SSIM). 

\paragraph{HCP Brains}
The Human Connectome Project dataset consists of 15 $128\times128\times128$ brain volumes, of which 7 volumes are used for training. The value of each voxel is a $6D$ vector representing the 6 free components of a $3\times3$ symmetric diffusion tensor (used to measure water diffusivity in the brain). The brains are separated into high and low resolution versions. The low resolution images were upsampled using $2\times$ nearest neighbour so the input and output equal in size. This is a good task to trial on, since superresolution in 3D is a memory intensive task. For training, we split the brain volumes into patches of size $24\times24\times24$ omitting patches with less than $1\%$ brain matter, resulting in an average of 112 patches per volume.

We evaluate on full brain volumes with the \textit{root mean squared error} (RMSE) between voxels containing brain matter in the ground-truth and the up-sampled volumes. We also calculate the error on the interior of the brain, defined by all voxels that are surrounded with a $5\times5$ cube within the full brain mask, to stay in line with prior literature \cite{tanno2017bayesian} \cite{blumberg2018deeper}.

\section{Results}\label{sec:results}
In this section, we evaluate the performance of our paired and unpaired RevGAN model, both quantitatively and qualitatively, against Pix2pix and CycleGAN baselines. Additionally, we study the scalability of our method in terms of memory-efficiency and model depth. For easy comparison, we  aim to use the same metrics as used in related literature. 

\begin{table}[b]
\centering
\resizebox{\linewidth}{!}{
\begin{tabular}{l c c c}
    \hline
    Model & RMSE (Interior) & RMSE (Total)\\
    \hline 
%    SRCNN $L_\text{L2}$ (baseline) & 7.14  $\pm$ 0.34  & 12.62 $\pm$ 0.52 \\
    Paired w/o $L_\text{GAN}$ (3D-SRCNN) & 7.03 $\pm$ 0.31 & 12.41 $\pm$ 0.57 \\
    Paired+2R w/o $L_\text{GAN}$ & 7.02 $\pm$ 0.32 & 12.41 $\pm$ 0.57 \\
    Paired+4R w/o $L_\text{GAN}$ & \textbf{6.68} $\pm$ 0.30 & \textbf{11.85} $\pm$ 0.56 \\
    Paired+8R w/o $L_\text{GAN}$ & 18.43 $\pm$ 1.03 & 21.40 $\pm$ 0.98 \\
    \hline    
    Paired (3D-Pix2pix)    & 11.94 $\pm$ 0.65  & 20.73 $\pm$ 1.05 \\
    Paired+2R              &  9.61 $\pm$ 0.40  & 17.36 $\pm$ 0.76 \\
    Paired+4R              &  8.43 $\pm$ 0.37  & 14.81 $\pm$ 0.61 \\
    Paired+8R              &  \textbf{7.82} $\pm$ 0.35  & \textbf{13.76} $\pm$ 0.60 \\
    \hline
    Unpaired (3D-CycleGAN) & 17.23 $\pm$ 0.73 & 26.94 $\pm$ 1.20 & \\
    Unpaired+2R            & \textbf{11.05} $\pm$ 0.51 & \textbf{17.76} $\pm$ 1.38 \\
    Unpaired+4R            & 18.98 $\pm$ 1.22 & 28.06 $\pm$ 1.44 \\
    Unpaired+8R            & 18.96 $\pm$ 0.85 & 27.94 $\pm$ 1.09 \\
\hline
\end{tabular}
}
\medskip
\caption{Mean and standard deviation of RMSE scores measured on the 8 brains in the \textit{HPC Brains} test set. Notice how in each experiment that the shallowest model is the not the highest performing. We are able to improve performance, by using deeper models at the same level of memory complexity as shallow models.}
\label{tab:results_3d}
\end{table}

\subsection{Qualitative Evaluation}
We present qualitative results of the RevGAN model on the \textit{Maps} dataset in Figure \ref{fig:maps_images} and on the \textit{Cityscapes} dataset in Figure \ref{fig:cityscapes_images}. We picked the first images in the dataset to avoid `cherry-picking' bias. The images are generated by models with equal parameter counts, indicated with a `$\dagger$' symbol in the quantitative results of the next section (Table \ref{tab:maps_results}, Table \ref{tab:cityscapes_results}).
% Depth 6 on the \texit{Cityscapes} dataset and depth 9 on the \textit{Maps} dataset, resulting in width 32 . The RevGAN model has width 56 at the same depth to obtain an equal parameter count of 3.9 M. For the \textit{Cityscape} dataset we used depth 9 and width 32 for Pix2pix and CycleGAN, following \cite{zhu2017unpaired}. For RevGAN we used width 58, resulting in a parameter count of 5.7 M ($\pm$ 0.1 M) for all models on this dataset.} in such a way that all models have the same amount of parameters for each dataset.

All models are able to produce images of similar or better visual quality. The greatest improvement can be seen in the unpaired model (compare CycleGAN with Unpaired RevGAN). Both paired tasks are visually more appealing than the unpaired tasks, which make intuitive sense, since paired image-to-image translation is an easier task to solve than the unpaired version. We therefore conclude that the RevGAN model does not seem to under-perform our non-reversible baselines in terms of observable visual quality. A more extensive collection of model outputs can be found in the supplementary material.

\begin{table*}[!t]
\centering
\resizebox{\linewidth}{!}{
\begin{tabular}{l r r l l l c l l l}
    \hline
\multicolumn{1}{c}{\multirow{2}{*}{\pbox{2cm}{Model}}} &
\multicolumn{1}{c}{\multirow{2}{*}{\pbox{2cm}{Width}}} &
\multicolumn{1}{c}{\multirow{2}{*}{\pbox{2cm}{Params}}}
& \multicolumn{3}{c}{maps$\rightarrow$satellite}
&
& \multicolumn{3}{c}{satellite$\rightarrow$maps}
\\
\cline{4-6} \cline{8-10}
& & & MAE & PSNR & SSIM & & MAE & PSNR & SSIM \\
\hline 
CycleGAN $^\dagger$   & 32 & 5.7 M
& 139.85 $\pm$ 15.52 & \textbf{14.62} $\pm$ 1.16 & 0.31 $\pm$ 0.05 &
& 138.86 $\pm$ 20.57 & \textbf{26.25} $\pm$ 3.64 & 0.81 $\pm$ 0.06 \\
Unpaired RevGAN & 32 & 1.7 M
& \textbf{133.57} $\pm$ 18.09 & 14.59 $\pm$ 0.96 & 0.31 $\pm$ 0.05 &
& 142.56 $\pm$ 18.94 & 26.23 $\pm$ 3.89 & 0.81 $\pm$ 0.06 \\
Unpaired RevGAN $^\dagger$& 58 & 5.6 M
& 134.63 $\pm$ 14.25 & 14.54 $\pm$ 1.09 & 0.30 $\pm$ 0.06 &  
& 148.98 $\pm$ 16.83 & 25.47 $\pm$ 4.27 & 0.80 $\pm$ 0.08 \\
Unpaired RevGAN & 64 & 6.8 M
& 135.48 $\pm$ 19.19 & 14.55 $\pm$ 1.24 & 0.26 $\pm$ 0.04 &
& \textbf{133.12} $\pm$ 17.18 & 23.66 $\pm$ 2.80 & 0.67 $\pm$ 0.10 \\
\hline
Pix2pix  $^\dagger$ & 32 & 5.7 M
& 139.63 $\pm$ 13.14 & 14.78 $\pm$ 1.08 & 0.30 $\pm$ 0.05 &
& 129.16 $\pm$ 16.11 & 27.11 $\pm$ 3.11 & 0.82 $\pm$ 0.04 \\
Paired RevGAN
& 32 & 1.7 M
& \textbf{139.23} $\pm$ 12.76 & 14.73 $\pm$ 1.07 & 0.30 $\pm$ 0.05 &
& 129.80 $\pm$ 15.54 & 26.84 $\pm$ 3.35 & 0.81 $\pm$ 0.05 \\
Paired RevGAN $^\dagger$ & 58 & 5.6 M
& 140.74 $\pm$ 12.45 & \textbf{14.91} $\pm$ 1.13 & 0.31 $\pm$ 0.05 &
& \textbf{128.55} $\pm$ 12.71 & 27.27 $\pm$ 3.12 & 0.82 $\pm$ 0.05 \\
Paired RevGAN & 64 & 6.8 M
& 140.59 $\pm$ 13.64 & 14.85 $\pm$ 1.20 & 0.31 $\pm$ 0.06 &
& 133.09 $\pm$ 12.09 & \textbf{27.37} $\pm$ 3.06 & 0.82 $\pm$ 0.04 \\
\hline
\end{tabular}
}
\medskip
\caption{Image quality on \textit{Maps} dataset. Notice how in most of the experiments that the RevGAN performs better than the baseline. $\dagger$ Parameter matched architectures}
\label{tab:maps_results}
\end{table*}

\subsection{Quantitative Evaluation}

\begin{figure*}[!b]
\resizebox{\linewidth}{!}{
\begin{tabular}{c | c | c c c c | c c c c r}
\multicolumn{2}{c}{} &
\multicolumn{4}{c}{\large Paired} &
\multicolumn{4}{c}{\large Unpaired} \\
\pbox[c]{\textwidth}{LR Input} &
\pbox[c]{\textwidth}{Ground-truth} &
\pbox[c]{\textwidth}{SRCNN \\ (3D-Pix2pix)} &
\pbox[c]{\textwidth}{RMSE} &
\pbox[c]{\textwidth}{\textbf{SRCNN+4R (Ours)}\\(Paired 3D-RevGAN)} &
\pbox[c]{\textwidth}{RMSE} &
\pbox[c]{\textwidth}{SRCNN \\ (3D-CycleGAN)} &
\pbox[c]{\textwidth}{RMSE} &
\pbox[c]{\textwidth}{\textbf{SRCNN+2R (Ours)}\\(Unpaired 3D-RevGAN)} &
\pbox[c]{\textwidth}{RMSE} & \\
\includegraphics[width=3.2cm]{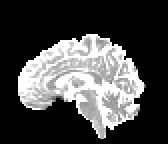} &
\includegraphics[width=3.2cm]{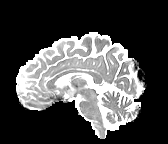} &
\includegraphics[width=3.2cm]{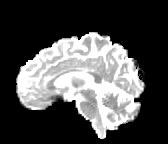} &
\includegraphics[width=3.2cm]{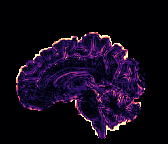} &
\includegraphics[width=3.2cm]{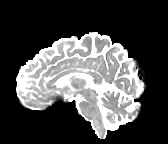} &
\includegraphics[width=3.2cm]{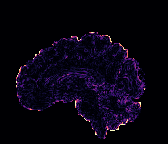} &
\includegraphics[width=3.2cm]{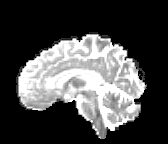} &
\includegraphics[width=3.2cm]{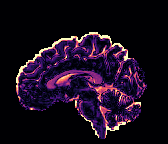} &
\includegraphics[width=3.2cm]{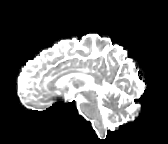} &
\includegraphics[width=3.2cm]{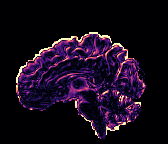} &
\multirow{2}[0]{*}[2.6cm]{\includegraphics[height=6.4cm]{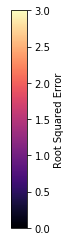}} 
\\

\includegraphics[width=3.2cm]{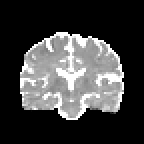} &
\includegraphics[width=3.2cm]{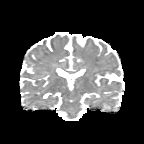} &
\includegraphics[width=3.2cm]{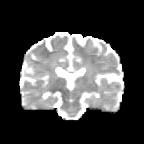} &
\includegraphics[width=3.2cm]{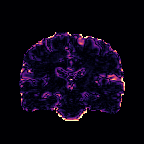} &
\includegraphics[width=3.2cm]{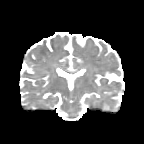} &
\includegraphics[width=3.2cm]{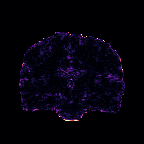} &
\includegraphics[width=3.2cm]{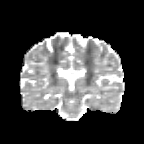} &
\includegraphics[width=3.2cm]{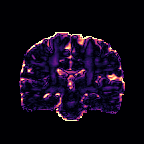} &
\includegraphics[width=3.2cm]{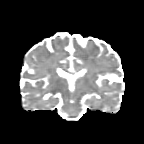} &
\includegraphics[width=3.2cm]{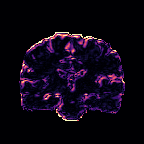} & \\
\end{tabular}
}
\medskip
\caption{Visualization of mean diffusivity maps on an sagittal slices (top) and axial slices (bottom) of the first brain in the \textit{HCP Brain} test set. From left to right: low-resolution input, high-resolution ground-truth, paired model without reversible layers (SRCNN-3D-Pix2pix), paired model with reversible layers (Paired 3D-RevGAN), unpaired model without reversible layers (SRCNN-3D-CycleGAN) and an unpaired model with reversible layers (Unpaired 3D-RevGAN).}
\label{fig:brain_images}
\end{figure*}

\paragraph{Cityscapes}
We provide quantitative evaluations of the performance of our RevGAN model on the \textit{Cityscapes} dataset. To ensure fairness, the baselines use the code implementations from the original papers. For our model, we provide two versions, a low parameter count version and a parameter matched version . In Table \ref{tab:cityscapes_results} the performance on the photo$\rightarrow$label mapping is given by segmentation scores in the center columns and the performance on the label$\rightarrow$photo is given by the \textit{FCN-Scores} in the righthand columns.

In Table \ref{tab:cityscapes_results}, we see that on the low parameter and parameter matched RevGAN models outperform the CycleGAN baselines on the per-pixel accuracy. This matches our qualitative observations from the previous section. For per-class and class IOU, we also beat the baseline on label$\rightarrow$photo, and from similar or marginally worse on the photo$\rightarrow$label task.

On the paired tasks we see that the results are more mixed and we perform roughly similar to the Pix2pix baseline, again matching our qualitative observations. We presume that the paired task is already fairly easy and thus the baseline performance is saturated. Thus introducing our model will do nothing to improve the visual quality of outputs. On the other hand, the unpaired task is harder and so the provision of by-design, approximately-inverse photo$\rightarrow$label and label$\rightarrow$photo generators improves visual quality. On the paired task, the main benefit comes in the form of the memory complexity (see Section \ref{sec:memory}), but on the unpaired task the RevGAN maintains low memory complexity, while generally improving numerical performance.

\paragraph{Maps}
Results on the \textit{Maps} dataset are shown in Table \ref{tab:maps_results}, which indicate that the RevGAN model performs similarly and sometimes better compared to the baselines. Again, similarly to the \emph{Cityscapes} experiment, we see that the biggest improvements are found on the unpaired tasks; whereas, the paired tasks demonstrate comparable performance.

\subsection{3D Volumes}
\label{sec:brains}

We also evaluate the performance of our RevGAN model on a 3D super-resolution problem, using the \textit{HTC Brains} dataset of \cite{tanno2017bayesian}. As baseline, we use a simple SRCNN model \cite{dong2014learning} (see supplementary material for architectural details) consisting of a $3\times3\times3$ convolutional layer as encoder, followed by a $1\times1\times1$ convolutional layer as decoder. For the discriminator, we use a 3D variant of the PatchGAN also used in \cite{zhu2017unpaired}. The RevGAN model, extends the architecture by inserting an invertible core between the encoder and the decoder. 

As can be seen in Figure \ref{fig:brain_images}, we obtain higher quality results using models with additional reversible residual layers. Of course, it is not unusual that deeper models result in higher quality predictions. Increasing the model size, however, is often unfeasible due to memory constraints. Fitting the activations in GPU memory can be particularly difficult when dealing with large 3D volumes. This study suggests that we can train deeper neural image-to-image translation models by adding reversible residual layers to existing architectures, without requiring more memory to store model activations.

\paragraph{With and without adversarial loss}
We performed the experiments on paired models with and without the adverarial loss $L_{\text{GAN}}$ . We found that models without such loss generally perform better in terms of pixel-distance, but that models with an adversarial loss typically obtain higher quality results upon visual inspection. A possible explanation of this phenomenon could be that models that solely minimize a pixel-wise distance, such as $L_1$ or $L_2$, tend to `average out' or blur the aleatoric uncertainty (natural diversity) that exists in the data, in order to obtain a low average loss. An adversarial loss enforces the model to output an image that could have been sampled from this uncertain distribution (thereby introduce realistic looking noise), often resulting in less blurry and visually more compelling renderings, but with a potentially higher pixel-wise error.

\subsection{Introspection}

\paragraph{Memory usage}
\label{sec:memory}
In this experiment, we evaluate the GPU memory consumption of our RevGAN model for increasing depths and compare it with a CycleGAN baseline. We kept the widths of both models fixed at such a value that the model parameters are approximately equal (both $\sim$3.9 M) at depth 6. 

As can be seen from Table \ref{tab:mem_usage}, the total memory usage increases for deeper networks in both models. In contrast to CycleGAN, however, the memory cost to store activations stays constant on the RevGAN model. A 6 layer CycleGAN model has the same total memory footprint of an unpaired RevGAN with 18-30 layers. Note that for convolutional layers the memory cost of storing the model is fixed given the network architecture, while the memory usage cost to store activations also depends on the size of the data. Therefore, reducing the memory cost of the activations becomes particularly important when training models on larger data sizes (e.g. higher image resolutions or increased batch sizes).

\begin{table}[t]
\centering
\begin{tabular}{c c l c c l}
    \hline
\multicolumn{1}{c}{\multirow{2}{*}{\pbox{2cm}{Depth}}} & \multicolumn{2}{c}{CycleGAN} & &  \multicolumn{2}{c}{Unpaired RevGAN} \\
% \multicolumn{1}{c}{} & \multicolumn{2}{c}{CycleGAN} & &  \multicolumn{2}{c}{RevGAN} \\
\cline{2-3} \cline{5-6}
& \small Model 
& \small Activations &
& \small Model
& \small Activations \\
%     Depth & Model & Activations & & Model & Activations \\
    \hline
% CycleGAN 32 vs RevGAN 64
%    6 &  1436.2 & + 746.6  & & 1384.0 & + 738.0\\
%    9 &  1462.0 & + 950.0  & & 1412.0 & + 738.0\\
%    30 & 1798.3 & + 2335.7 & & 1576.0 & + 738.0\\
%    60 & 2278.3 & + 4315.6 & & 1814.0 & + 738.0\\
% CycleGAN 32 vs RevGAN 56 (more fair comparison)
%     6 &   574.3 & + 752.0  & &  512.4 & + \textbf{646.1} \\
%      9 &   622.3 & + 949.9  & &  526.5 & + \textbf{646.1} \\
%     30 &  958.8 & + 2335.9 & &  610.4 & + \textbf{646.1} \\
%     60 & 1279.5 & + 4315.6 & &  689.4 & + \textbf{646.1} \\
6 & 434.3 & + 752.0 & & 374.4 & + \textbf{646.1} \\
9 & 482.3 & + 949.0 & & 385.4 & + \textbf{646.1} \\
12 & 530.3 & + 1148.1 & & 398.5 & + \textbf{646.1} \\
18 & 626.3 & + 1543.9 & & 423.4 & + \textbf{646.1} \\
30 & 818.7 & + 2335.8 & & 626.3 & + \textbf{646.1} \\
\hline
\end{tabular}
\medskip 
\caption{Memory usage on GPU measured in MiB on a single Nvidia Tesla K40m GPU on the \textit{Maps} dataset (lower is better). Both the CycleGAN and unpaired RevGAN have a similar number of parameters.}
\label{tab:mem_usage}
\end{table}

\begin{figure}[b]
\begin{center}
\resizebox{\linewidth}{!}{
\subfloat[Per-pixel accuracy against depth.]{
\includegraphics[height=3.3cm]{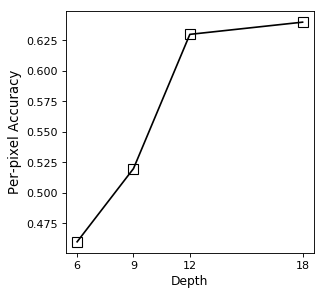}
}
\hspace{0.3cm}
\subfloat[Memory usage against depth.]{
\includegraphics[height=3.3cm]{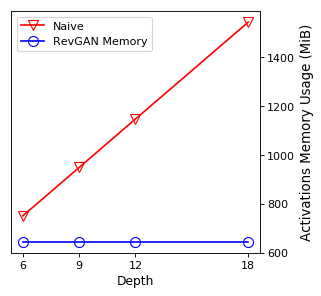}
}
}
\end{center}
\caption{Comparison of per-pixel accuracy for a width 64 RevGAN evaluated after 75 epochs and memory usage on \textit{Cityscapes} dataset.}
\label{fig:acc_mem}
\end{figure}

\paragraph{Scalability} Reversible architectures can be trained arbitarily deep without increasing the memory cost needed to store activations. We evaluate the performance of larger RevGAN models on the \textit{Cityscapes} dataset.

% \begin{table}[H]
% \begin{center}
% \begin{tabular}{l r p{1.3cm} p{1.3cm} p{0.6cm}}
%     \hline
%     Depth & Params & Per-pixel acc. & Per-class acc. & Class IOU\\ \hline
%     6 & 5.1 M & 0.46 & 0.20 & 0.12 \\
%     9 & 6.8 M & 0.52 & 0.22 & 0.15 \\
%     12 & 8.6 M & 0.63 & 0.23 & 0.17 \\
%     18 & 12.1 M & 0.64 & 0.23 & 0.17 \\ \hline
% \end{tabular}
% \end{center}
% \caption{Deeper networks: Classification-scores on \textit{Cityscapes} photo$\rightarrow$label.}
% \label{tab:cityscapes_abl1}
% \end{table}
% 
% \begin{table}[H]
% \begin{center}
% \begin{tabular}{l r p{1.3cm} p{1.3cm} p{0.6cm}}
%     \hline
%     Depth & Params & Per-pixel acc. & Per-class acc. & Class IOU\\ \hline
%     6  & 5.1 M & 0.27 & 0.16 & 0.03 \\
%     9  & 6.8 M & 0.39 & 0.14 & 0.10 \\
%     12 & 8.6 M & 0.60 & 0.21 & 0.15 \\
%     18 & 12.1 M & 0.59 & 0.20 & 0.15 \\
%     \hline
% \end{tabular}
% \end{center}
% \caption{Deeper networks: FCN-scores on \textit{Cityscapes} label$\rightarrow$photo.}
% \label{tab:cityscapes_abl2}
% \end{table}

As shown in Figure \ref{fig:acc_mem}, with successive increases in depth, the performance of the RevGAN model increases on the \textit{Cityscapes} task. This effect seems to hold up until a certain depth ($\sim 12-18$) after which we find a slight decrease in performance again. We presume this decrease in performance is due to the longer training times of deeper models, which we have not been able to train to full convergence due to time-budgeting issues. Keep in mind that we tried to keep our network architectures and training parameters as close as possible to networks used in the original Pix2pix and CycleGAN models. Other research suggests that training models with much deeper reversible architectures can be very effective \cite{chang2017reversible}. We leave the exploration of alternative reversible architectures to future work.

\section{Limitations and Discussion}
Our results indicate that we can train image-to-image translation models with close to constant memory requirements in depth (see Table \ref{tab:mem_usage}). This enables us to scale up to very deep architectures. Our ablation studies also show that increasing depth can lead to higher quantitative results in terms of various semantic segmentation metrics. This ability to scale up, however, trades memory for time, and so there is a trade-off to be considered in practical situations where we may be concerned about how long to spend in the development phase of such models. This is evident in our ablation study in Figure \ref{fig:acc_mem}, where we were not able to wait until full convergence of the deepest models.

We have also demonstrated empirically that given a constrained budget of trainable parameters, we are able to achieve improved performance on the Cityscapes and Maps datasets, especially for of an unpaired training regime. We accredit two mechanisms for this observation. 
% The first is that we need not learn a separate encoder and decoder, so training the $X\to Y$ mapping 

Due to the nature of the problem, our network is not fully invertible. As a result, we still need to use the cycle-consistency loss, which requires two forward propagation passes and two backward passes through the model. A possible way to circumvent using the the cycle-consistency loss is to design the encoders and decoders to be analytically pseudo-invertible. We in fact did experiments on this, by formulating the (strided-)convolutions as Toeplitz matrix–vector products \cite{matuson1985svd}. Unfortunately, we found that exact pseudo-invertibility is computationally too slow to run. Another issue with our setup is that two discriminators are required during training time (one of each domain). These are not used at test time, and can thus be considered as superfluous networks, requiring a lot of extra memory. That said, this is a general problem with CycleGAN and Pix2pix models in general.

\section{Conclusion}
In this paper we have proposed a new image-to-image translation model using reversible residual layers. The proposed model is approximately invertible \emph{by design}, essentially weight-tying in the forward and backward direction, hence training from domain $X$ to domain $Y$ simulaneously trains the mapping from $Y$ to $X$. We demonstrate equivalent or improved performance in terms of image quality, compared to similar non-reversible methods. Additionally, we show that our model is more memory efficient, because activations of reversible residual layers do not have to be stored to perform backpropagation.

In future work we plan to explore techniques to get rid of the cycle-consistency loss, so that the network is automatically cycle-consistent to begin with. 

%\section*{Acknowledgments}
%Authors would like to thank Bram van Ginneken for valuable discussion and feedback. We are also grateful to the Diagnostic Image Analysis Group (DIAG) and the Netherlands Organisation for Scientific Research (NWO) for supporting this research and providing computational resources.

\begin{figure*}[h]
\resizebox{\linewidth}{!}{
\begin{tabular}{c | c c | c c | c}
Input & CycleGAN & \textbf{Unpaired RevGAN (ours)} & Pix2pix & \textbf{Paired RevGAN (ours)} & Ground-truth  \\
\includegraphics[height=1.9cm,width=3.8cm]{images/cityscapes_images/1_city_cycle_real_A.png}&
\includegraphics[height=1.9cm,width=3.8cm]{images/cityscapes_images/1_city_cycle_fake_B.png} &
\includegraphics[height=1.9cm,width=3.8cm]{images/cityscapes_images/1_city_core_fake_B.png} &
\includegraphics[height=1.9cm,width=3.8cm]{images/cityscapes_images/1_city_pix2pix_fake_B.png} &
\includegraphics[height=1.9cm,width=3.8cm]{images/cityscapes_images/1_city_pix3pix_fake_B.png} &
\includegraphics[height=1.9cm,width=3.8cm]{images/cityscapes_images/1_city_cycle_real_B.png} \\ 
\includegraphics[height=1.9cm,width=3.8cm]{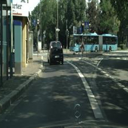}&
\includegraphics[height=1.9cm,width=3.8cm]{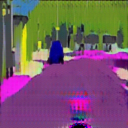} &
\includegraphics[height=1.9cm,width=3.8cm]{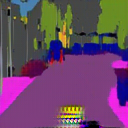} &
\includegraphics[height=1.9cm,width=3.8cm]{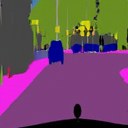} &
\includegraphics[height=1.9cm,width=3.8cm]{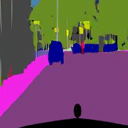} &
\includegraphics[height=1.9cm,width=3.8cm]{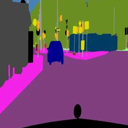} \\ 
\includegraphics[height=1.9cm,width=3.8cm]{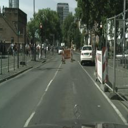}&
\includegraphics[height=1.9cm,width=3.8cm]{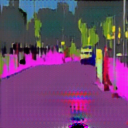} &
\includegraphics[height=1.9cm,width=3.8cm]{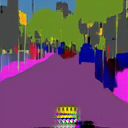} &
\includegraphics[height=1.9cm,width=3.8cm]{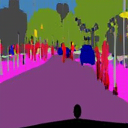} &
\includegraphics[height=1.9cm,width=3.8cm]{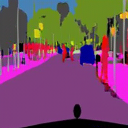} &
\includegraphics[height=1.9cm,width=3.8cm]{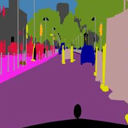} \\ 
\includegraphics[height=1.9cm,width=3.8cm]{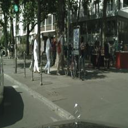}&
\includegraphics[height=1.9cm,width=3.8cm]{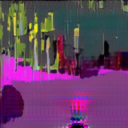} &
\includegraphics[height=1.9cm,width=3.8cm]{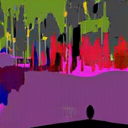} &
\includegraphics[height=1.9cm,width=3.8cm]{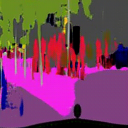} &
\includegraphics[height=1.9cm,width=3.8cm]{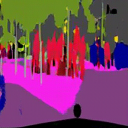} &
\includegraphics[height=1.9cm,width=3.8cm]{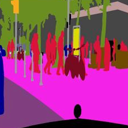} \\ 
\includegraphics[height=1.9cm,width=3.8cm]{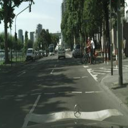}&
\includegraphics[height=1.9cm,width=3.8cm]{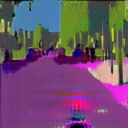} &
\includegraphics[height=1.9cm,width=3.8cm]{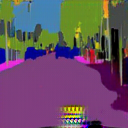} &
\includegraphics[height=1.9cm,width=3.8cm]{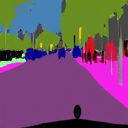} &
\includegraphics[height=1.9cm,width=3.8cm]{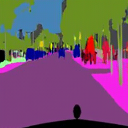} &
\includegraphics[height=1.9cm,width=3.8cm]{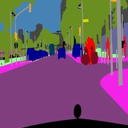} \\ 
\includegraphics[height=1.9cm,width=3.8cm]{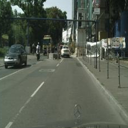}&
\includegraphics[height=1.9cm,width=3.8cm]{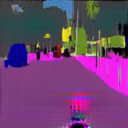} &
\includegraphics[height=1.9cm,width=3.8cm]{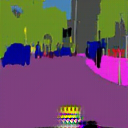} &
\includegraphics[height=1.9cm,width=3.8cm]{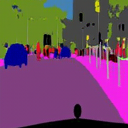} &
\includegraphics[height=1.9cm,width=3.8cm]{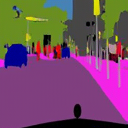} &
\includegraphics[height=1.9cm,width=3.8cm]{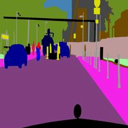} \\ 
\includegraphics[height=1.9cm,width=3.8cm]{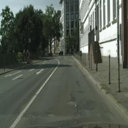}&
\includegraphics[height=1.9cm,width=3.8cm]{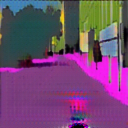} &
\includegraphics[height=1.9cm,width=3.8cm]{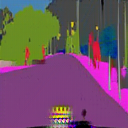} &
\includegraphics[height=1.9cm,width=3.8cm]{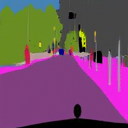} &
\includegraphics[height=1.9cm,width=3.8cm]{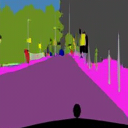} &
\includegraphics[height=1.9cm,width=3.8cm]{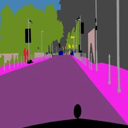} \\ 
\end{tabular}
}
\medskip
\caption{Additional image mappings for \textit{photo$\rightarrow$label} on the \textit{Cityscapes} test set.}
\label{fig:more_cityscapes_images_a}
\end{figure*}
\begin{figure*}[h]
\resizebox{\linewidth}{!}{
\begin{tabular}{c | c c | c c | c}
Input & CycleGAN & \textbf{Unpaired RevGAN (ours)} & Pix2pix & \textbf{Paired RevGAN (ours)} & Ground-truth  \\
\includegraphics[height=1.9cm,width=3.8cm]{images/cityscapes_images/1_city_cycle_real_B.png} &
\includegraphics[height=1.9cm,width=3.8cm]{images/cityscapes_images/1_city_cycle_fake_A.png} &
\includegraphics[height=1.9cm,width=3.8cm]{images/cityscapes_images/1_city_core_fake_A.png} &
\includegraphics[height=1.9cm,width=3.8cm]{images/cityscapes_images/1_city_pix2pix_fake_A.png} &
\includegraphics[height=1.9cm,width=3.8cm]{images/cityscapes_images/1_city_pix3pix_fake_A.png} &
\includegraphics[height=1.9cm,width=3.8cm]{images/cityscapes_images/1_city_cycle_real_A.png} \\
\includegraphics[height=1.9cm,width=3.8cm]{images/cityscapes_images/10_city_cycle_real_B.png} &
\includegraphics[height=1.9cm,width=3.8cm]{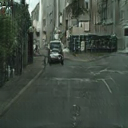} &
\includegraphics[height=1.9cm,width=3.8cm]{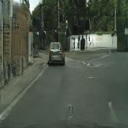} &
\includegraphics[height=1.9cm,width=3.8cm]{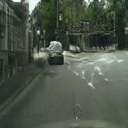} &
\includegraphics[height=1.9cm,width=3.8cm]{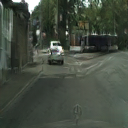} &
\includegraphics[height=1.9cm,width=3.8cm]{images/cityscapes_images/10_city_cycle_real_A.png} \\
\includegraphics[height=1.9cm,width=3.8cm]{images/cityscapes_images/100_city_cycle_real_B.png} &
\includegraphics[height=1.9cm,width=3.8cm]{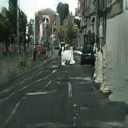} &
\includegraphics[height=1.9cm,width=3.8cm]{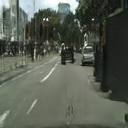} &
\includegraphics[height=1.9cm,width=3.8cm]{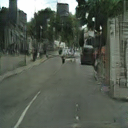} &
\includegraphics[height=1.9cm,width=3.8cm]{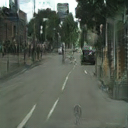} &
\includegraphics[height=1.9cm,width=3.8cm]{images/cityscapes_images/100_city_cycle_real_A.png} \\
\includegraphics[height=1.9cm,width=3.8cm]{images/cityscapes_images/101_city_cycle_real_B.png} &
\includegraphics[height=1.9cm,width=3.8cm]{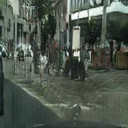} &
\includegraphics[height=1.9cm,width=3.8cm]{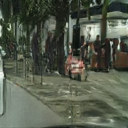} &
\includegraphics[height=1.9cm,width=3.8cm]{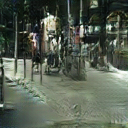} &
\includegraphics[height=1.9cm,width=3.8cm]{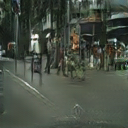} &
\includegraphics[height=1.9cm,width=3.8cm]{images/cityscapes_images/101_city_cycle_real_A.png} \\
\includegraphics[height=1.9cm,width=3.8cm]{images/cityscapes_images/102_city_cycle_real_B.png} &
\includegraphics[height=1.9cm,width=3.8cm]{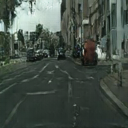} &
\includegraphics[height=1.9cm,width=3.8cm]{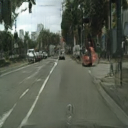} &
\includegraphics[height=1.9cm,width=3.8cm]{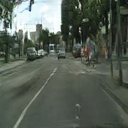} &
\includegraphics[height=1.9cm,width=3.8cm]{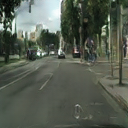} &
\includegraphics[height=1.9cm,width=3.8cm]{images/cityscapes_images/102_city_pix3pix_fake_A.png} \\
\includegraphics[height=1.9cm,width=3.8cm]{images/cityscapes_images/103_city_cycle_real_B.png} &
\includegraphics[height=1.9cm,width=3.8cm]{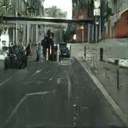} &
\includegraphics[height=1.9cm,width=3.8cm]{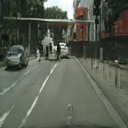} &
\includegraphics[height=1.9cm,width=3.8cm]{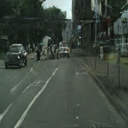} &
\includegraphics[height=1.9cm,width=3.8cm]{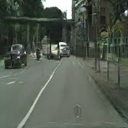} &
\includegraphics[height=1.9cm,width=3.8cm]{images/cityscapes_images/103_city_cycle_real_A.png} \\
\includegraphics[height=1.9cm,width=3.8cm]{images/cityscapes_images/104_city_cycle_real_B.png} &
\includegraphics[height=1.9cm,width=3.8cm]{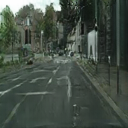} &
\includegraphics[height=1.9cm,width=3.8cm]{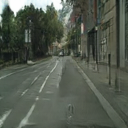} &
\includegraphics[height=1.9cm,width=3.8cm]{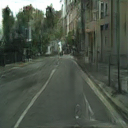} &
\includegraphics[height=1.9cm,width=3.8cm]{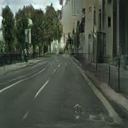} &
\includegraphics[height=1.9cm,width=3.8cm]{images/cityscapes_images/104_city_cycle_real_A.png} \\
\end{tabular}
}
\medskip
\caption{Additional image mappings for \textit{label$\rightarrow$photo} on the \textit{Cityscapes} test set.}
\label{fig:more_cityscapes_images_b}
\end{figure*}

\begin{figure*}[p]
\resizebox{\linewidth}{!}{
\begin{tabular}{c | c c | c c | c}
Input & CycleGAN & \textbf{Unpaired RevGAN (ours)} & Pix2pix & \textbf{Paired RevGAN (ours)} & Ground-truth \\
\includegraphics[width=0.17\linewidth]{images/maps_images_jpg/1_maps_cycle_real_A.jpg} &
\includegraphics[width=0.17\linewidth]{images/maps_images_jpg/1_maps_cycle_fake_B.jpg} &
\includegraphics[width=0.17\linewidth]{images/maps_images_jpg/1_maps_core_fake_B.jpg} &
\includegraphics[width=0.17\linewidth]{images/maps_images_jpg/1_maps_pix2pix_fake_B.jpg} &
\includegraphics[width=0.17\linewidth]{images/maps_images_jpg/1_maps_pix3pix_fake_B.jpg} &
\includegraphics[width=0.17\linewidth]{images/maps_images_jpg/1_maps_cycle_real_B.jpg} \\
\includegraphics[width=0.17\linewidth]{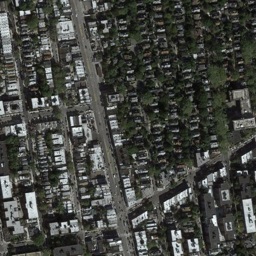} &
\includegraphics[width=0.17\linewidth]{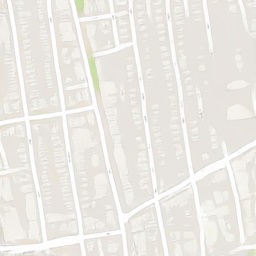} &
\includegraphics[width=0.17\linewidth]{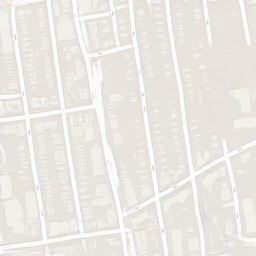} &
\includegraphics[width=0.17\linewidth]{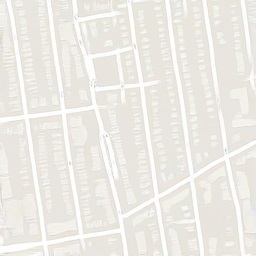} &
\includegraphics[width=0.17\linewidth]{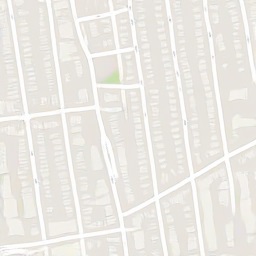} &
\includegraphics[width=0.17\linewidth]{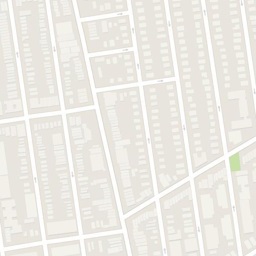} \\
\includegraphics[width=0.17\linewidth]{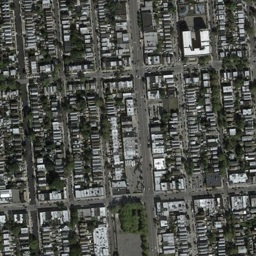} &
\includegraphics[width=0.17\linewidth]{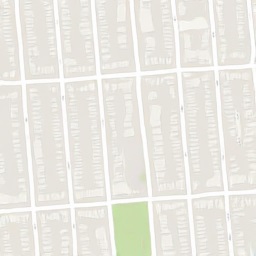} &
\includegraphics[width=0.17\linewidth]{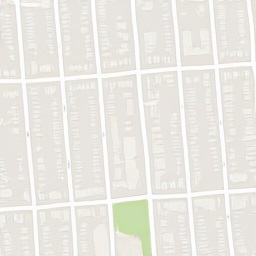} &
\includegraphics[width=0.17\linewidth]{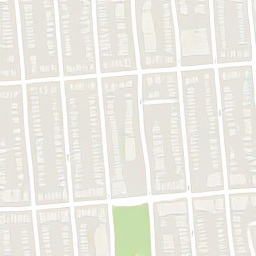} &
\includegraphics[width=0.17\linewidth]{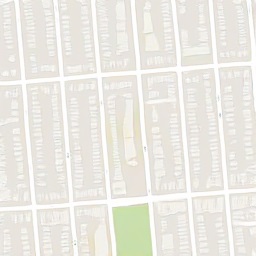} &
\includegraphics[width=0.17\linewidth]{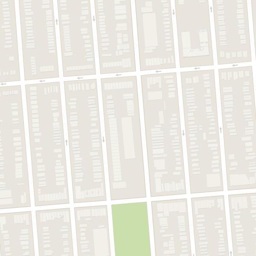} \\
\includegraphics[width=0.17\linewidth]{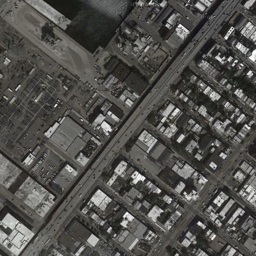} &
\includegraphics[width=0.17\linewidth]{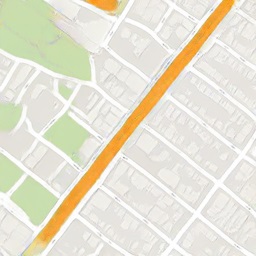} &
\includegraphics[width=0.17\linewidth]{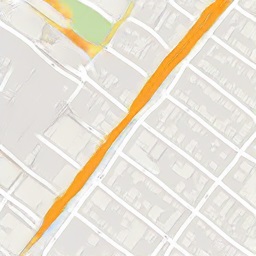} &
\includegraphics[width=0.17\linewidth]{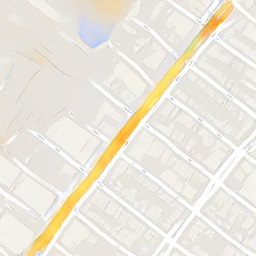} &
\includegraphics[width=0.17\linewidth]{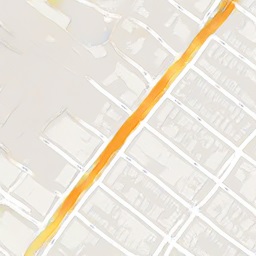} &
\includegraphics[width=0.17\linewidth]{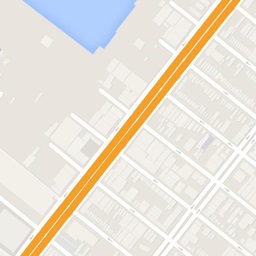} \\
\includegraphics[width=0.17\linewidth]{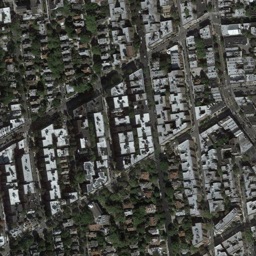} &
\includegraphics[width=0.17\linewidth]{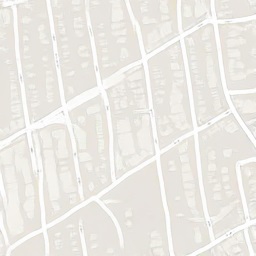} &
\includegraphics[width=0.17\linewidth]{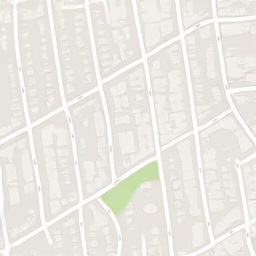} &
\includegraphics[width=0.17\linewidth]{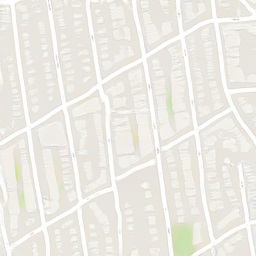} &
\includegraphics[width=0.17\linewidth]{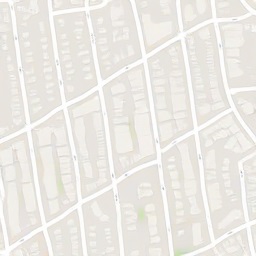} &
\includegraphics[width=0.17\linewidth]{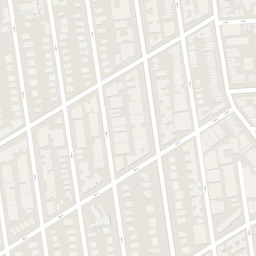} \\
\includegraphics[width=0.17\linewidth]{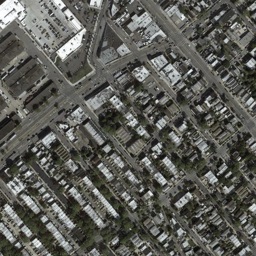} &
\includegraphics[width=0.17\linewidth]{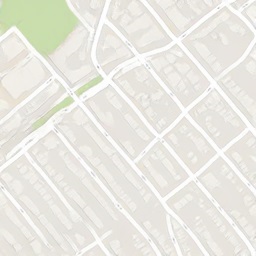} &
\includegraphics[width=0.17\linewidth]{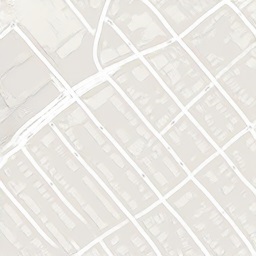} &
\includegraphics[width=0.17\linewidth]{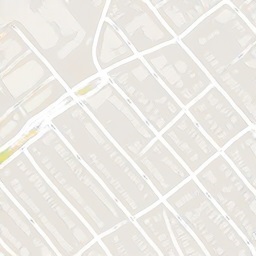} &
\includegraphics[width=0.17\linewidth]{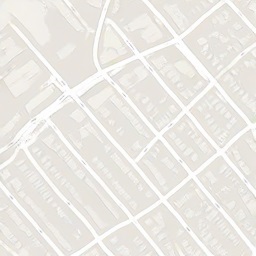} &
\includegraphics[width=0.17\linewidth]{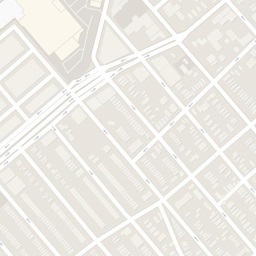} \\
\includegraphics[width=0.17\linewidth]{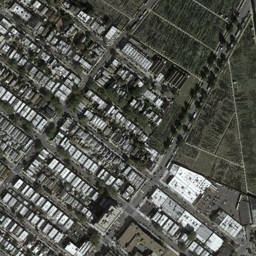} &
\includegraphics[width=0.17\linewidth]{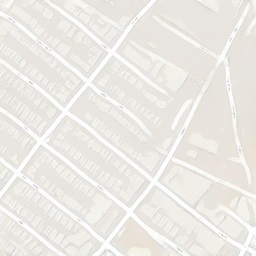} &
\includegraphics[width=0.17\linewidth]{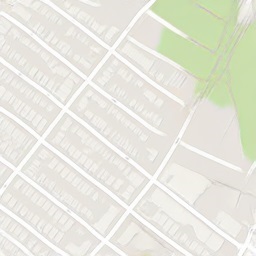} &
\includegraphics[width=0.17\linewidth]{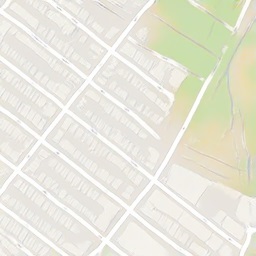} &
\includegraphics[width=0.17\linewidth]{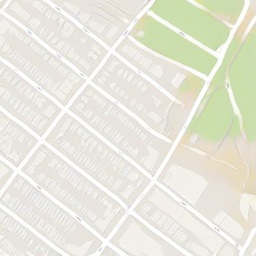} &
\includegraphics[width=0.17\linewidth]{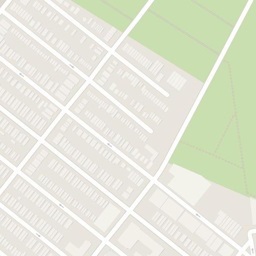} \\
\includegraphics[width=0.17\linewidth]{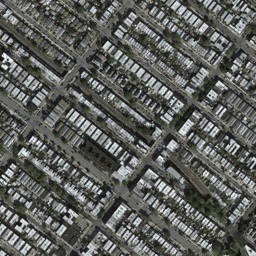} &
\includegraphics[width=0.17\linewidth]{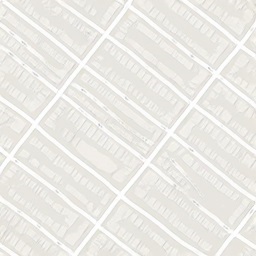} &
\includegraphics[width=0.17\linewidth]{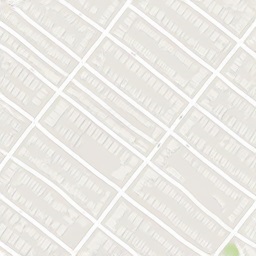} &
\includegraphics[width=0.17\linewidth]{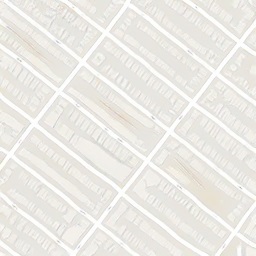} &
\includegraphics[width=0.17\linewidth]{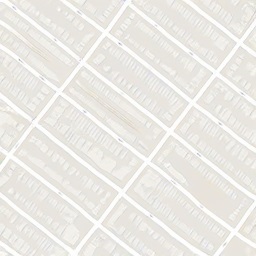} &
\includegraphics[width=0.17\linewidth]{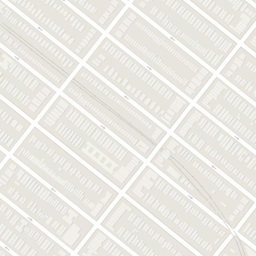} \\
\end{tabular}
}
\medskip
\caption{Additional image mappings for \textit{satellite$\rightarrow$maps} on \textit{Maps} test set.}
\label{fig:more_maps_images_a}
\end{figure*}

\begin{figure*}[p]
\resizebox{\linewidth}{!}{
\begin{tabular}{c | c c | c c | c}
Input & CycleGAN & \textbf{Unpaired RevGAN (ours)} & Pix2pix & \textbf{Paired RevGAN (ours)} & Ground-truth \\
\includegraphics[width=0.17\linewidth]{images/maps_images_jpg/1_maps_cycle_real_B.jpg} &
\includegraphics[width=0.17\linewidth]{images/maps_images_jpg/1_maps_cycle_fake_A.jpg} &
\includegraphics[width=0.17\linewidth]{images/maps_images_jpg/1_maps_core_fake_A.jpg} &
\includegraphics[width=0.17\linewidth]{images/maps_images_jpg/1_maps_pix2pix_fake_A.jpg} &
\includegraphics[width=0.17\linewidth]{images/maps_images_jpg/1_maps_pix3pix_fake_A.jpg} &
\includegraphics[width=0.17\linewidth]{images/maps_images_jpg/1_maps_cycle_real_A.jpg}\\
\includegraphics[width=0.17\linewidth]{images/maps_images_jpg/10_maps_cycle_real_B.jpg} &
\includegraphics[width=0.17\linewidth]{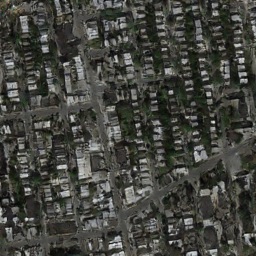} &
\includegraphics[width=0.17\linewidth]{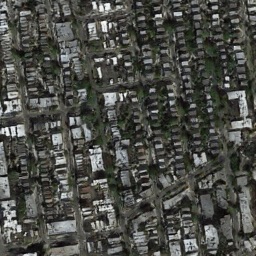} &
\includegraphics[width=0.17\linewidth]{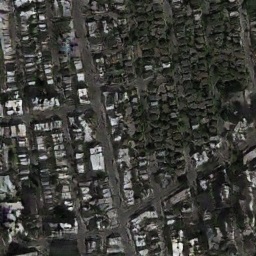} &
\includegraphics[width=0.17\linewidth]{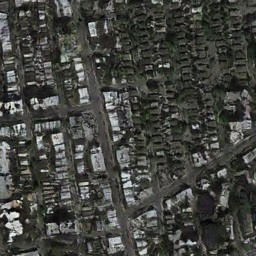} &
\includegraphics[width=0.17\linewidth]{images/maps_images_jpg/10_maps_cycle_real_A.jpg}\\
\includegraphics[width=0.17\linewidth]{images/maps_images_jpg/100_maps_cycle_real_B.jpg} &
\includegraphics[width=0.17\linewidth]{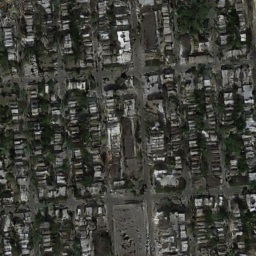} &
\includegraphics[width=0.17\linewidth]{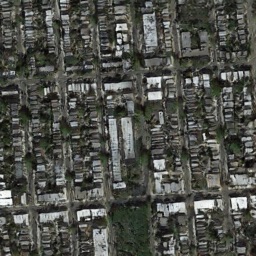} &
\includegraphics[width=0.17\linewidth]{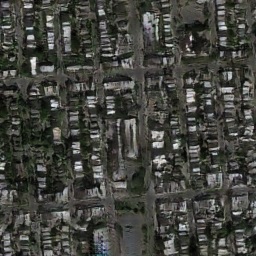} &
\includegraphics[width=0.17\linewidth]{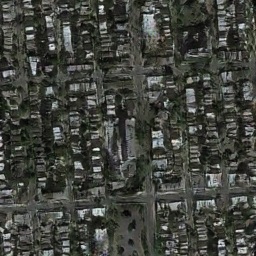} &
\includegraphics[width=0.17\linewidth]{images/maps_images_jpg/100_maps_cycle_real_A.jpg}\\
\includegraphics[width=0.17\linewidth]{images/maps_images_jpg/1000_maps_cycle_real_B.jpg} &
\includegraphics[width=0.17\linewidth]{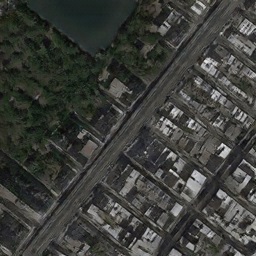} &
\includegraphics[width=0.17\linewidth]{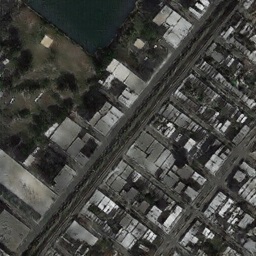} &
\includegraphics[width=0.17\linewidth]{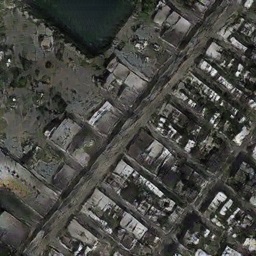} &
\includegraphics[width=0.17\linewidth]{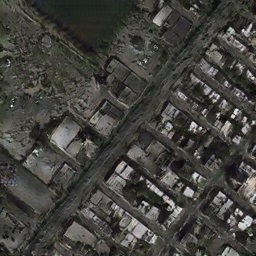} &
\includegraphics[width=0.17\linewidth]{images/maps_images_jpg/1000_maps_cycle_real_A.jpg}\\
\includegraphics[width=0.17\linewidth]{images/maps_images_jpg/1001_maps_cycle_real_B.jpg} &
\includegraphics[width=0.17\linewidth]{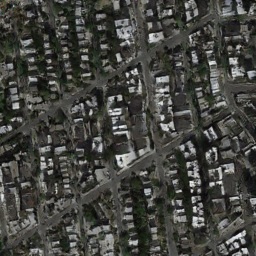} &
\includegraphics[width=0.17\linewidth]{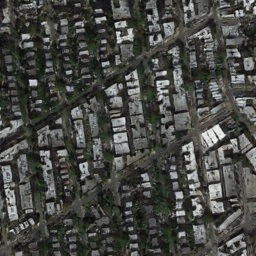} &
\includegraphics[width=0.17\linewidth]{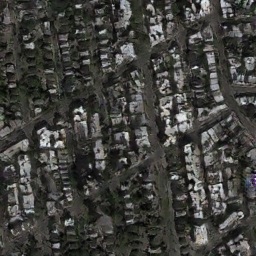} &
\includegraphics[width=0.17\linewidth]{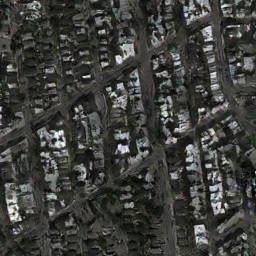} &
\includegraphics[width=0.17\linewidth]{images/maps_images_jpg/1001_maps_cycle_real_A.jpg}\\
\includegraphics[width=0.17\linewidth]{images/maps_images_jpg/1002_maps_cycle_real_B.jpg} &
\includegraphics[width=0.17\linewidth]{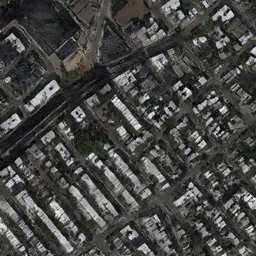} &
\includegraphics[width=0.17\linewidth]{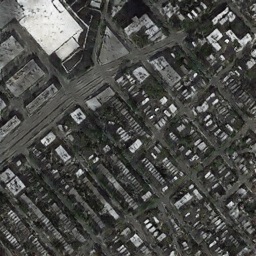} &
\includegraphics[width=0.17\linewidth]{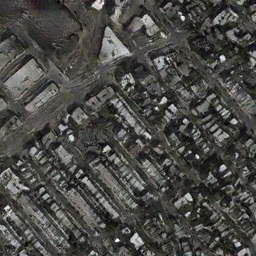} &
\includegraphics[width=0.17\linewidth]{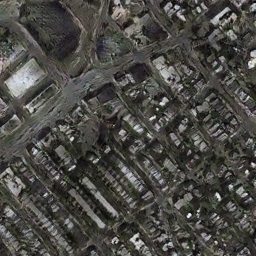} &
\includegraphics[width=0.17\linewidth]{images/maps_images_jpg/1002_maps_cycle_real_A.jpg}\\
\includegraphics[width=0.17\linewidth]{images/maps_images_jpg/1003_maps_cycle_real_B.jpg} &
\includegraphics[width=0.17\linewidth]{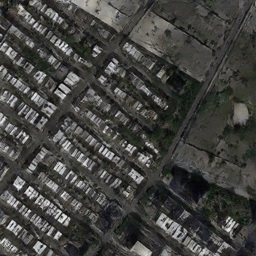} &
\includegraphics[width=0.17\linewidth]{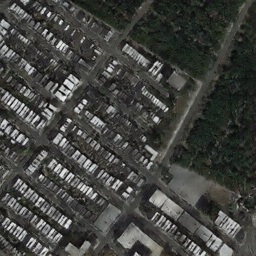} &
\includegraphics[width=0.17\linewidth]{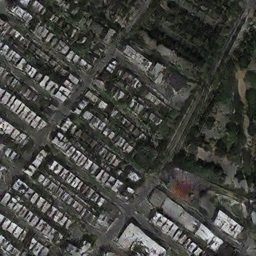} &
\includegraphics[width=0.17\linewidth]{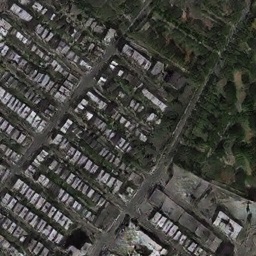} &
\includegraphics[width=0.17\linewidth]{images/maps_images_jpg/1003_maps_cycle_real_A.jpg}\\
\includegraphics[width=0.17\linewidth]{images/maps_images_jpg/1004_maps_cycle_real_B.jpg} &
\includegraphics[width=0.17\linewidth]{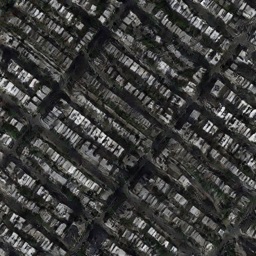} &
\includegraphics[width=0.17\linewidth]{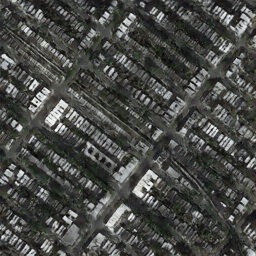} &
\includegraphics[width=0.17\linewidth]{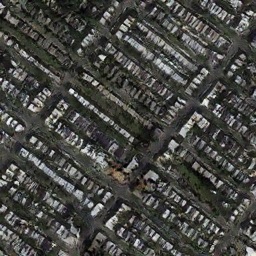} &
\includegraphics[width=0.17\linewidth]{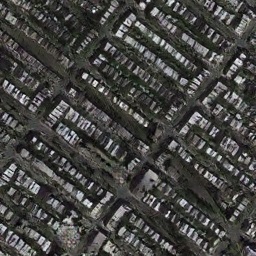} &
\includegraphics[width=0.17\linewidth]{images/maps_images_jpg/1004_maps_cycle_real_A.jpg}\\
\end{tabular}
}
\medskip
\caption{Additional image mappings for \textit{maps$\rightarrow$satellite} on \textit{Maps} test set.}
\label{fig:more_maps_images_b}
\end{figure*}

\clearpage
%{\small
%\bibliographystyle{ieee}
%\bibliography{egbib}
%}

\clearpage
\section*{Appendix A: Implementation Details}

We provide a Pytorch \cite{paszke2017automatic} implementation on Github. Our code extends the image-to-image translation framework from \cite{zhu2017unpaired} with several reversible models in 2D and 3D. The reversible blocks are implemented using a modified version of MemCNN \cite{van2018memcnn}.

\subsection{Generator architecture}
\tikzstyle{empty} = [circle]
\tikzstyle{b_image} = [rectangle, minimum width=3.5cm, minimum height=1cm, text centered, draw=black, fill=white!20]
\tikzstyle{b_sample} = [rectangle, minimum width=3.5cm, minimum height=1cm, text centered, draw=black, fill=orange!50]
\tikzstyle{b_conv} = [rectangle, minimum width=3.5cm, minimum height=1cm, text centered, draw=black, fill=green!5!blue!20]
\tikzstyle{b_norm} = [rectangle, minimum width=3.5cm, minimum height=1cm, text centered, draw=black, fill=orange!50]
\tikzstyle{b_rev} = [rectangle, minimum width=3.5cm, minimum height=1cm, text centered, draw=black, fill=red!30]
\tikzstyle{b_func} = [rectangle, minimum width=3.5cm, minimum height=1cm, text centered, draw=black, fill=black!10!green!30]
\tikzstyle{arrow} = [thick,-Triangle,>=stealth]

\paragraph{2d Architecture}
All 2d models adapt network architectures similar to those used in \cite{zhu2017unpaired} and \cite{johnson2016perceptual}. The encoders $\text{Enc}_X, \text{Dec}_Y$ consist of a $7\times7$ convolutional layer that maps 3 input channels to $K$ channels, followed by two $3\times3$ convolutional layers with stride 2 that spatially downsample ($/4$) the signal and increase ($\times2$) the channel dimension. We also refer to $K$ as the \textit{width} of our network. As reversible core $C$, we use $R$ sequential reversible residual layers (with $R=6$ for $128\times128$ \textit{Cityscapes} data and $R=9$ for $256\times256$ \textit{Maps} data). We consider the amount of reversible residual layers in the core to be the \textit{depth} of our network. The decoders $\text{Dec}_X$ and $\text{Dec}_Y$ are build out of two $3\times3$ fractionally-strided convolutional layers \footnote{`Fractionally-strided convolutional layers' or `transposed convolutions' are sometimes referred to as `deconvolutions' in literature. To avoid confusion, especially in the context of invertibility, we follow this \cite{dumoulin2016guide} guide on convolutional arithmetic, and only refer to the term 'deconvolution' when we speak of the mathematical inverse of a convolution, which is different from the fractionally-strided convolution.}, followed by a $7\times7$ convolutional layer projecting the final features to 3 output channels.

We apply reflection padding before every convolution to avoid spatial downsampling. Each convolutional layer is followed by an instance normalization layer \cite{ulyanovinstance} and a ReLU nonlinearity, except for the last convolutional layer which is directly followed by a Tanh non-linearity to scale the output within $[-1, 1]$, just like the normalized data.

\begin{figure}[H]
\centering
\resizebox{0.5\linewidth}{!}{
\begin{tikzpicture}[node distance=1.7cm]
\node (start) [b_image] {Input Image};
\node (a) [b_conv, below of=start] {$7\times7$ Conv};
\node (b) [b_conv, below of=a] {$3\times3$ Conv (stride 2)};
\node (c) [b_conv, below of=b] {$3\times3$ Conv (stride 2)};
\node (d) [b_rev, below of=c] {2D Reversible Block};
\node (d_label) [empty, xshift=1cm, right of=d] {\Large $\times$ R};
\node (e) [b_conv, below of=d] {$3\times3$ Transposed Conv};
\node (f) [b_conv, below of=e] {$3\times3$ Transposed Conv};
\node (g) [b_conv, below of=f] {$7\times7$ Conv};
\node (end) [b_image, below of=g] {Output Image};
\draw [arrow] (start) -- node[anchor=west] {$3\times W\times H$} (a);
\draw [arrow] (a) -- node[anchor=west] {$C\times W\times H$} (b);
\draw [arrow] (b) -- node[anchor=west] {$2C\times \frac{1}{2}W\times \frac{1}{2}H$} (c);
\draw [arrow] (c) -- node[anchor=west] {$4C\times \frac{1}{4}W\times \frac{1}{4}H$} (d);
\draw [arrow] (d) -- node[anchor=west] {$4C\times \frac{1}{4}W\times \frac{1}{4}H$} (e);
\draw [arrow] (e) -- node[anchor=west] {$2C\times \frac{1}{2}W\times \frac{1}{2}H$} (f);
\draw [arrow] (f) -- node[anchor=west] {$C\times W\times W$} (g);
\draw [arrow] (g) -- node[anchor=west] {$3\times W\times H$} (end);
\draw [decorate,decoration={brace,amplitude=5pt},xshift=0pt,yshift=-200pt]
(-2.5,1.2) -- (-2.5,6.0) node [black,midway,xshift=-20pt,thick] {\footnotesize
$\text{Enc}$};
\draw [decorate,decoration={brace,amplitude=5pt},xshift=0pt,yshift=-200pt]
(-2.5,-0.5) -- (-2.5,0.8) node [black,midway,xshift=-20pt,thick] {\footnotesize
$\text{Core}$};
\draw [decorate,decoration={brace,amplitude=5pt},xshift=0pt,yshift=-200pt]
(-2.5,-5.6) -- (-2.5,-0.8) node [black,midway,xshift=-20pt,thick] {\footnotesize
$\text{Dec}$};
\node[text width=3.3cm, align=right, font=\scriptsize] at (0,-2) {+ Instance norm + ReLU};
\node[text width=3.3cm, align=right, font=\scriptsize] at (0,-3.7) {+ Instance norm + ReLU};
\node[text width=3.3cm, align=right, font=\scriptsize] at (0,-5.4) {+ Instance norm + ReLU};
\node[text width=3.3cm, align=right, font=\scriptsize] at (0,-7.1) {(See Figure 2)};
\node[text width=3.3cm, align=right, font=\scriptsize] at (0,-8.8) {+ Instance norm + ReLU};
\node[text width=3.3cm, align=right, font=\scriptsize] at (0,-10.5) {+ Instance norm + ReLU};
\node[text width=3.3cm, align=right, font=\scriptsize] at (0,-12.2) {+ Tanh};
\end{tikzpicture}
}
\caption{2D Generator Architecture}
\label{fig:2d_generator}
\end{figure}
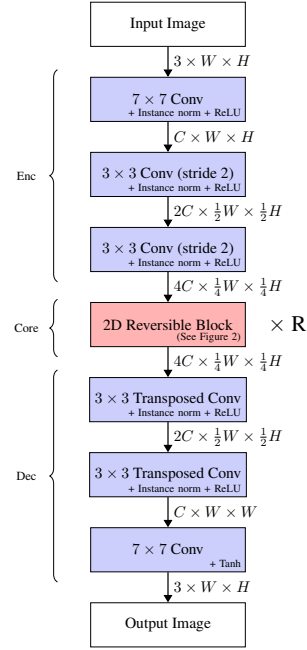

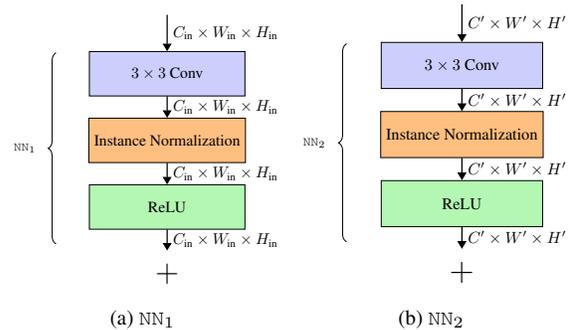
\begin{figure}[b]
\centering
\subfloat[$\texttt{NN}_1$]{
\resizebox{0.45\linewidth}{!}{
\centering
\begin{tikzpicture}[node distance=1.5cm]
\node (start) [empty] {};
\node (a) [b_conv, below of=start] {$3\times3$ Conv};
\node (b) [b_norm, below of=a] {Instance Normalization};
\node (c) [b_func, below of=b] {ReLU};
\node (end) [empty, below of=c] {\huge $+$};
\draw [arrow] (start) -- node[anchor=west] {$C_{\text{in}}\times W_{\text{in}}\times H_{\text{in}}$} (a);
\draw [arrow] (a) -- node[anchor=west] {$C_{\text{in}}\times W_{\text{in}}\times H_{\text{in}}$} (b);
\draw [arrow] (b) -- node[anchor=west] {$C_{\text{in}}\times W_{\text{in}}\times H_{\text{in}}$} (c);
\draw [arrow] (c) -- node[anchor=west] {$C_{\text{in}}\times W_{\text{in}}\times H_{\text{in}}$} (end);
\draw [decorate,decoration={brace,amplitude=5pt},xshift=0pt,yshift=-200pt]
(-2.5,1.7) -- (-2.5,6.0) node [black,midway,xshift=-20pt,thick] {\footnotesize
$\texttt{NN}_1$};
\end{tikzpicture}
}}
\subfloat[$\texttt{NN}_2$]{
\resizebox{0.45\linewidth}{!}{
\centering
\begin{tikzpicture}[node distance=1.5cm]
\node (start) [empty] {};
\node (a) [b_conv, below of=start] {$3\times3$ Conv};
\node (b) [b_norm, below of=a] {Instance Normalization};
\node (c) [b_func, below of=b] {ReLU};
\node (end) [empty, below of=c] {\huge $+$};
\draw [arrow] (start) -- node[anchor=west] {$C'\times W'\times H'$} (a);
\draw [arrow] (a) -- node[anchor=west] {$C'\times W'\times H'$} (b);
\draw [arrow] (b) -- node[anchor=west] {$C'\times W'\times H'$} (c);
\draw [arrow] (c) -- node[anchor=west] {$C'\times W'\times H'$} (end);
\draw [decorate,decoration={brace,amplitude=5pt},xshift=0pt,yshift=-200pt]
(-2.5,1.7) -- (-2.5,6.0) node [black,midway,xshift=-20pt,thick] {\footnotesize
$\texttt{NN}_2$};
\end{tikzpicture}
}}
\caption{Schematic representation of $\texttt{NN}_1$ and $\texttt{NN}_2$ in 2D Reversible Residual Block.}
\label{fig:2d_revblocks}
\end{figure}

A full schematic version of the 2D architecture can be found in Figure \ref{fig:2d_generator}. A diagram of the (identical) $\texttt{NN}_1$ and $\texttt{NN}_2$ functions used in the 2D reversible block are shown in Figure \ref{fig:2d_revblocks}.

\paragraph{3d Architecture}
For the 3-dimensional super-resolution task (\textit{HTC Brains}), we consider our input and output to be equally sized. Therefore, we first up-sample the images from the  low-resolution input domain, before feeding them to the model. It is known that this method also helps to prevent checkerboard-like artifacts \cite{odena2016deconvolution}. The first layer in our model is a $3\times3\times3$ convolution layer that increases the channel dimension to $K$, and is directly followed by an instance normalization layer and a ReLU non-linearity. Then we apply an arbitrary amount of 3D reversible blocks using additive coupling, with the following sequence for $\texttt{NN}_1$ and $\texttt{NN}_2$: a $3\times3\times3$ convolutional layer, an instance normalization layer, a ReLU non-linearity and another $3\times3\times3$ convolution. We use reflection padding of 1 to ensure that the $\texttt{NN}_1$ and $\texttt{NN}_2$ are volume-preserving. Also, we initialize the reversible blocks perform as the identity mapping, by initializing the weights of the last convolutional layer in the reversible block with zeros. This trick has previously shown to be effective in the context of reversible networks \cite{kingma2018glow}.

A full schematic version of the 3D generator can be found in Figure \ref{fig:3d_generator}. A diagram illustrating $\texttt{NN}_1$ and $\texttt{NN}_2$ used in the 3D reversible block is shown in Figure \ref{fig:3d_revblocks}.

\begin{figure}[H]
\centering
\resizebox{0.5\linewidth}{!}{
\begin{tikzpicture}[node distance=1.5cm]
\node (start) [b_image] {Input Image};
\node (a) [b_conv, below of=start] {$3\times3\times3$ Conv};
\node (b) [b_rev, below of=a] {3D Reversible Block};
\node (b_label) [empty, xshift=1cm, right of=b] {\Large $\times$ R};
\node (c) [b_conv, below of=b] {$1\times1\times1$ Conv};
\node (end) [b_image, below of=c] {Output Image};
\draw [arrow] (start) -- node[anchor=west] {$6\times24\times24$} (a);
\draw [arrow] (a) -- node[anchor=west] {$K\times24\times24$} (b);
\draw [arrow] (b) -- node[anchor=west] {$K\times24\times24$} (c);
\draw [arrow] (c) -- node[anchor=west] {$6\times24\times24$} (end);
\node[text width=3.3cm, align=right, font=\scriptsize] at (0,-1.80) {+ Instance norm + ReLU};
\node[text width=3.3cm, align=right, font=\scriptsize] at (0,-3.35) {(See Figure 4)};
\node[text width=3.3cm, align=right, font=\scriptsize] at (0,-4.8) {+ Tanh};
\draw [decorate,decoration={brace,amplitude=5pt},xshift=0pt,yshift=-200pt]
(-2.5,5) -- (-2.5,6) node [black,midway,xshift=-20pt,thick] {\footnotesize
$\text{Enc}$};
\draw [decorate,decoration={brace,amplitude=5pt},xshift=0pt,yshift=-200pt]
(-2.5,3.5) -- (-2.5,4.5) node [black,midway,xshift=-20pt,thick] {\footnotesize
$\text{Core}$};
\draw [decorate,decoration={brace,amplitude=5pt},xshift=0pt,yshift=-200pt]
(-2.5,2.0) -- (-2.5,3.0) node [black,midway,xshift=-20pt,thick] {\footnotesize
$\text{Dec}$};
\end{tikzpicture}
}
\caption{3D RevGAN Architecture}
\label{fig:3d_generator}
\end{figure}

\tikzset{fontscale/.style = {font=\relsize{#1}}}
\tikzstyle{stateTransition}=[-stealth, thick]

\begin{figure}[H]
\centering
\subfloat[]{
\resizebox{0.45\linewidth}{!}{
\begin{tikzpicture}[node distance=1.5cm]
\node (start) [empty] {};
\node (a) [b_conv, below of=start] {$3\times3\times3$ Conv};
\node (b) [b_norm, below of=a] {Instance Normalization};
\node (c) [b_func, below of=b] {ReLU};
\node (d) [b_conv, below of=c] {$3\times3\times3$ Conv (0 init)};
\node (end) [empty, below of=d] {\huge $+$};
\draw [arrow] (start) -- node[anchor=west] {$C' \times W'\times H'$} (a);
\draw [arrow] (a) -- node[anchor=west] {$C' \times W'\times H'$} (b);
\draw [arrow] (b) -- node[anchor=west] {$C' \times W'\times H'$} (c);
\draw [arrow] (c) -- node[anchor=west] {$C' \times W'\times H'$} (d);
\draw [arrow] (d) -- node[anchor=west] {$C' \times W'\times H'$} (end);
\draw [decorate,decoration={brace,amplitude=5pt},xshift=0pt,yshift=-200pt]
(-2.5,-0.8) -- (-2.5,6) node [black,midway,xshift=-20pt] {\footnotesize
$\text{NN}_1$};
\end{tikzpicture}
}}
\subfloat[]{
\resizebox{0.45\linewidth}{!}{
\begin{tikzpicture}[node distance=1.5cm]
\node (start) [empty] {};
\node (a) [b_conv, below of=start] {$3\times3\times3$ Conv};
\node (b) [b_norm, below of=a] {Instance Normalization};
\node (c) [b_func, below of=b] {ReLU};
\node (d) [b_conv, below of=c] {$3\times3\times3$ Conv (0 init)};
\node (end) [empty, below of=d] {\huge $+$};
\draw [arrow] (start) -- node[anchor=west] {$C' \times W'\times H'$} (a);
\draw [arrow] (a) -- node[anchor=west] {$C' \times W'\times H'$} (b);
\draw [arrow] (b) -- node[anchor=west] {$C' \times W'\times H'$} (c);
\draw [arrow] (c) -- node[anchor=west] {$C' \times W'\times H'$} (d);
\draw [arrow] (d) -- node[anchor=west] {$C' \times W'\times H'$} (end);
\draw [decorate,decoration={brace,amplitude=5pt},xshift=0pt,yshift=-200pt]
(-2.5,-0.8) -- (-2.5,6) node [black,midway,xshift=-20pt,thick] {\footnotesize
$\text{NN}_2$};
\end{tikzpicture}
}}

\caption{3D RevGAN Architecture}
\label{fig:3d_revblocks}
\end{figure}

\subsection{Discriminator Architecture}
For the discriminator, we adapt the same architecture as used in \cite{zhu2017unpaired}, also known as PatchGAN. We use subsequent $4\times4$ convolutional layers with stride 2 followed by LeakyReLU (with 0.2 slope) non-linearities. The first layer projects the input to 64 layers, followed by three layers each doubling the channel dimension. Finally, we obtain a 1-dimensional outputs by applying a $1\times1$ convolution followed by a Sigmoid.
The 3D models use a very similar architecture and solely replacing the 2D convolutional kernels by equally sized 3D convolutional layers (e.g. $3\times3$ kernels become $3\times3\times3$ kernels).

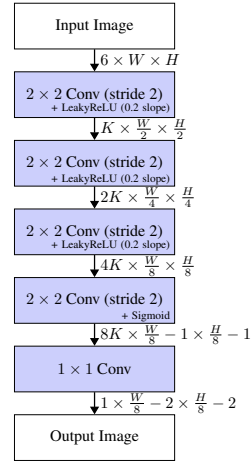
\begin{figure}[H]
\centering
\resizebox{0.4\linewidth}{!}{
\begin{tikzpicture}[node distance=1.5cm]
\node (start) [b_image] {Input Image};
\node (a) [b_conv, below of=start] {$2\times2$ Conv (stride 2)};
\node (b) [b_conv, below of=a] {$2\times2$ Conv (stride 2)};
\node (c) [b_conv, below of=b] {$2\times2$ Conv (stride 2)};
\node (d) [b_conv, below of=c] {$2\times2$ Conv (stride 2)};
\node (e) [b_conv, below of=d] {$1\times1$ Conv};
\node (end) [b_image, below of=e] {Output Image};
\draw [arrow] (start) -- node[anchor=west] {$6\times W\times H$} (a);
\draw [arrow] (a) -- node[anchor=west] {$K\times \frac{W}{2}\times \frac{H}{2}$} (b);
\draw [arrow] (b) -- node[anchor=west] {$2K\times \frac{W}{4}\times \frac{H}{4}$} (c);
\draw [arrow] (c) -- node[anchor=west] {$4K\times \frac{W}{8}\times \frac{H}{8}$} (d);
\draw [arrow] (d) -- node[anchor=west] {$8K\times \frac{W}{8}-1\times \frac{H}{8}-1$} (e);
\draw [arrow] (e) -- node[anchor=west] {$1\times \frac{W}{8}-2\times \frac{H}{8}-2$} (end);
\node[text width=3.3cm, align=right, font=\scriptsize] at (0,-1.80) {+ LeakyReLU (0.2 slope)};
\node[text width=3.3cm, align=right, font=\scriptsize] at (0,-3.35) {+ LeakyReLU (0.2 slope)};
\node[text width=3.3cm, align=right, font=\scriptsize] at (0,-4.8) {+ LeakyReLU (0.2 slope)};
\node[text width=3.3cm, align=right, font=\scriptsize] at (0,-6.35) {+ Sigmoid};
\end{tikzpicture}
}
\caption{2D PatchGAN Discriminator}
\label{fig:2d_discriminator}
\end{figure}

\subsection{Hyper-parameters}
A summary of the used hyper-parameters can be found in Table \ref{tab:params} below.

\begin{table}[H]
\centering
\resizebox{0.4\textwidth}{!}{
\begin{tabular}{|c|c|c|}
\hline
\textbf{Parameter} & \textbf{2D} & \textbf{3D} \\ 
\hline
Data size & \pbox{3cm}{$3\times128\times128$ or \\$3\times256\times256$} & $6\times24\times24\times24$ \\ \hline
Weight initialization & \multicolumn{2}{c|}{$\mathcal{N}(\mu=0, \sigma=0.02)$} \\ \hline
Normalization & \multicolumn{2}{c|}{Instance Norm} \\\hline
Dropout & \multicolumn{2}{c|}{No} \\ \hline
Optimizer & \multicolumn{2}{c|}{Adam \cite{kingma2014adam}} \\ \hline
Optimizer params & \multicolumn{2}{c|}{$\beta_1=0.5, \beta_2=0.999$} \\ \hline
Epochs & 200 & 20 \\ \hline
Batch size & \multicolumn{2}{c|}{1} \\ \hline
Learning rate & \multicolumn{2}{c|}{0.002} \\ \hline
Learning rate decay & \multicolumn{2}{c|}{\pbox{6cm}{Keep fixed first half of epochs. \\ Linearly decay to 0 in second half of epochs.}} \\ \hline
\end{tabular}
}
\caption{Summary of hyper-parameters}
\label{tab:params}
\end{table}

\clearpage
\section*{Appendix B: Negative Results}
\begin{itemize}
\item We tried to replace \textit{additive coupling} with \textit{affine coupling}, which has been applied succesfully in the context of reversible networks by \cite{kingma2018glow}. In theory, affine coupling layers are more general and more expressive than additive coupling. We found, however, that affine coupling degraded performance and made training more unstable. Nevertheless, it would be interesting to see whether affine coupling outperforms additive coupling combined with other architectures or hyper-parameters.
 
\item We tried to replace the down-sampling and up-sampling layers with sub-pixel convolutions \cite{shi2016real} in our 2D and 3D models, which have also been applied succesfully in the context of invertible architectures \cite{jacobsen2018revnet}, but found that it degraded performance. Sub-pixel convolutions were originally proposed to save memory in super-resolution problems by applying convolutions in lower-dimensional space rather than in the higher-dimensional target space. The RevGAN model, on the other hand, saves memory by not having to store the activations of the reversible layers.

\item We found that the invertible core can be replaced with a continuous-depth residual networks introduced in \cite{chen2018neural} of which the forward and inverse pass are trained using an ordinary differential equation (ODE) solver. The method has some practical benefits, including constant $\mathcal{O}(1)$ memory cost as a function of depth, similar to reversible residual layers, and explicit control over the numerical error. Due to time constraints, we were not able to compare the method with residual layers. It would be interesting to explore the use of neural ordinary (or even stochastic) differential equations in the context of image-to-image translation.

\item  We tried \textit{Consensus Optimization} \cite{mescheder2017numerics} to stabilize training by encouraging agreement between the discriminators and the generators. Consensus optimization boils down to regularization term over the second-order derivative over our gradients, which is a computationally intensive task. We stopped using it because it slowed down training too much.

\item We tried to replace the transposed convolutions used for up-sampling in our model with nearest-neighbour and bilinear upsampling to prevent checkerboard-like aftifacts as explained in \cite{odena2016deconvolution}, but found that it degraded performance. Furthermore, we observed that the checkerboard appeared in early training stages, but that they disappeared after a sufficient amount of training iterations.

\end{itemize}

\end{document}